\documentclass[lettersize,journal]{IEEEtran}
\usepackage{amsmath,amsfonts}
\usepackage{algorithmic}
\usepackage{algorithm}
\usepackage{array}
\usepackage[caption=false,font=normalsize,labelfont=sf,textfont=sf]{subfig}
\usepackage{textcomp}
\usepackage{stfloats}
\usepackage{url}
\usepackage{verbatim}
\usepackage{graphicx}
\usepackage{cite}
\usepackage{xcolor} 
\usepackage{booktabs}
\usepackage{tabularx}
\usepackage{multirow}
\usepackage{multicol}
\usepackage{hyperref} 
\hypersetup{
colorlinks=true,
linkcolor=blue,
citecolor=blue
}

\begin{document}

\title{CV-Cities: Advancing Cross-View Geo-Localization in Global Cities}

\author{Gaoshuang Huang, Yang Zhou*, Luying Zhao, and Wenjian Gan
        
\thanks{This manuscript was accepted by IEEE JSTARS. \textit{(Corresponding author Yang Zhou)}}

\thanks{The authors are with the Institute of Geospatial Information, Information Engineering University, Zheng Zhou 450001, China (e-mail: huanggaoshuang123@163.com; zhouyang3d@163.com).}}

\markboth{arXiv}%
{Shell \MakeLowercase{\textit{et al.}}: A Sample Article Using IEEEtran.cls for IEEE Journals}


\maketitle

\begin{abstract}
Cross-view geo-localization (CVGL), which involves matching and retrieving satellite images to determine the geographic location of a ground image, is crucial in GNSS-constrained scenarios. However, this task faces significant challenges due to substantial viewpoint discrepancies, the complexity of localization scenarios, and the need for global localization. To address these issues, we propose a novel CVGL framework that integrates the vision foundational model DINOv2 with an advanced feature mixer. Our framework introduces the symmetric InfoNCE loss and incorporates near-neighbor sampling and dynamic similarity sampling strategies, significantly enhancing localization accuracy. Experimental results show that our framework surpasses existing methods across multiple public and self-built datasets. To further improve global-scale performance, we have developed CV-Cities, a novel dataset for global CVGL. CV-Cities includes $223,736$ ground-satellite image pairs with geolocation data, spanning sixteen cities across six continents and covering a wide range of complex scenarios, providing a challenging benchmark for CVGL. The framework trained with CV-Cities demonstrates high localization accuracy in various test cities, highlighting its strong globalization and generalization capabilities. Our datasets and codes are available at \url{https://github.com/GaoShuang98/CVCities}.
\end{abstract}


\section{Introduction}
\IEEEPARstart{I}{n} complex environments such as forests and urban canyons, GNSS signals are often weakened or lost due to factors such as tree shading, building shading, building reflections, and electromagnetic interference, severely compromising positioning accuracy\cite{ref1, ref2, ref3, ref4}. Cross-view geo-localization (CVGL) is a new positioning technique that does not rely on GNSS signals but speculates the location by comparing the similarity between ground and pre-acquired satellite images. Since satellite images can be conveniently acquired globally, this method is widely applicable globally and shows significant advantages in urban canyon user localization, automated driving, event detection, military intelligence acquisition, drone navigation, disaster response, etc.

Nevertheless, despite the recent advancements in CVGL techniques, research in this field still presents some challenges. The effective extraction and matching of image features with different viewpoints represent a complex problem, particularly in environments characterized by dense buildings, seasonal changes, or rapidly developing urban landscapes\cite{ref5}. Furthermore, enhancing the robustness and accuracy of CVGL to accommodate diverse and intricate global scenes represents a pivotal avenue of current research.

The rapid development of vision foundational models has made generating generalized visual features from images possible. Due to the training using hundreds of millions of data, the vision foundational model can extract image features that are more general and robust than those extracted by traditional models. This enables it to cope effectively with the huge viewpoint differences in CVGL. This paper proposes a new framework for CVGL. It employs the vision foundational model DINOv2\cite{ref6} with a feature mix module\cite{ref7} as a model to extract robust global features. The effective symmetric InfoNCE loss\cite{ref8, ref9, ref10} is used for training, and near-neighbor sampling and dynamic similarity sampling strategies are employed to improve the model’s accuracy.

Most existing CVGL datasets are limited to a specific region, with a relatively narrow range of scenes and limited data. This constrains the model’s global generalization performance. This paper presents CV-Cities (Cross-View geo-localization in global Cities), a global CVGL dataset, to improve the model’s generalization performance. The results of the experiments show that the CV-Cities markedly enhance the model’s generalization ability. Following testing in a variety of cities across the globe, the model showed high geo-localization accuracy. Furthermore, the test set included in CV-Cities encompasses a multitude of intricate scenarios, including city, natural, seasonal change, water area, occlusion, and more. This represents a novel and formidable benchmark. Our key contributions are as follows:
\begin{enumerate}
    \item We have established a global CVGL dataset, CV-Cities, which offers the most comprehensive coverage and diverse range of images available for CVGL. The dataset facilitates enhanced geo-localization accuracy and generalization performance for the model. CV-Cities encompass a range of complex scenarios, and the test and training set do not intersect, thereby providing a highly challenging benchmark for CVGL.
    
    \item We propose a novel CVGL framework employing a model with shared weight parameters to extract features from two viewpoints images. This framework does not require the processing of images, such as polar coordinate conversion or image generation, which results in a simple structure and higher accuracy. Furthermore, the framework’s performance has been evaluated, resulting in the highest CVGL accuracy to date.

    \item Two sampling strategies, near neighbor sampling (NNS) and dynamic similarity sampling (DSS), were employed to mine negative samples for model training. This approach improves the accuracy of the framework’s localization.

    \item A comparison was conducted between the CV-Cities and other datasets, demonstrating that the CV-Cities offers a significant advantage in improving model generalization.
\end{enumerate}

\section{Related Works}
\subsection{CVGL methods}
In the early stages of CVGL, manual features were employed to facilitate the matching process between ground and satellite images. For instance, Lin et al.\cite{ref11} initially employed manual features, including HOG\cite{ref12}, Gist\cite{ref13}, and color histogram, for CVGL. Viswanathan\cite{ref14} transformed ground query images into satellite images and matched them using manual features, such as SIFT\cite{ref15}, SURF\cite{ref16}, and FREAK\cite{ref17}. Although manual features are simple and interpretable, they struggle with viewpoint discrepancies, resulting in low matching accuracy.

With the rise of deep learning, CNN-based methods became dominant. Lin et al.\cite{ref18} introduced Where-CNN, achieving $80\%$ top $20\%$ accuracy, but highlighting the need for improvement in practical applications. Workman et al.\cite{ref19} enhanced geo-localization by optimizing the CNN architecture and introducing the CVUSA dataset, now a key benchmark in CVGL. Tian et al.\cite{ref20} applied Faster R-CNN\cite{ref21} for object detection, followed by twin neural networks for retrieval and a dominance set-based matching method. Hu et al.\cite{ref22} developed CVM-Net, which uses convolutional layers with NetVLAD\cite{ref23} encoding to extract global features. More recently, Deuser et al.\cite{ref8} proposed Sample4Geo, utilizing ConvNeXt with symmetric InfoNCE loss, demonstrating high performance across multiple datasets. Li et al.\cite{ref23-1} used CNN as the backbone, and proposed a framework called patch similarity self-knowledge distillation. Despite these advancements, challenges in accuracy and efficiency persist, as CNN-based methods like Faster R-CNN are computationally expensive, and models like CVM-Net and Sample4Geo still face generalization issues across datasets.

The introduction of Vision Transformers (ViT) has significantly improved cross-view matching accuracy by leveraging self-attention mechanisms. Dai et al.\cite{ref24} introduced a feature segmentation and region alignment structure to enhance contextual understanding. Yang et al.\cite{ref25} utilized ViT’s self-attention to model global dependencies, reducing visual variability. Other works, like Zhuang et al.\cite{ref26}, integrated pixel-based attention for UAV-to-remote sensing image matching, and Wang et al.\cite{ref27} combined CNN and transformer networks to improve feature embedding. Zhang et al.\cite{ref28} proposed the GeoDTR model to separate geometric information from visual features using a transformer-based extractor. Despite their accuracy, ViT-based models are computationally intensive and prone to overfitting on small datasets, limiting scalability.

Geometric transformations and image generation methods have also been explored to minimize viewpoint differences. For example, Shi et al.\cite{ref2} used polar coordinate transformations to convert satellite images into panoramic views, while Zhao et al.\cite{ref29} employed geometric projections to transform ground panoramic images into satellite viewpoints. Regmi et al.\cite{ref30} and Hao et al.\cite{ref31} utilized GANs for cross-view image conversion, with some methods achieving high fidelity but also introducing artifacts that can degrade accuracy. Toker et al.\cite{ref32} proposed a GAN-based retrieval method, and Huang et al.\cite{ref33} introduced a sequential fork network for generating ground images from satellite views. However, these methods face limitations, as GAN-generated images often suffer from artifacts, and geometric transformations only partially reduce viewpoint discrepancies. Moreover, the GAN generation step can also reduce the overall efficiency of the CVGL process.

\subsection{CVGL Datasets}
The CVGL dataset utilized for training and testing is summarized in Table \ref{tab1}. CVUSA\cite{ref19} comprises ground and satellite images sourced from various regions within the United States. The ground images are derived from Google Street View, with a resolution of $1,232 \times 224$ pixels, while the satellite images are obtained from Microsoft Bing Maps, with a zoom level of $19$ and a resolution of $750 \times 750$ pixels. The dataset encompasses $35,532$ and $8,884$ pairs of images for training and testing, respectively. It represents the inaugural dataset to be used in a cross-view matching study. The ground images included in this dataset have been aligned with the orientation of the satellite images and thus can be used as an additional feature.

CVACT\cite{ref34} provides a collection of Canberra ground and satellite images. The ground panoramic images are captured from Google Street View, with a resolution of $1,664 \times 832$ pixels. The satellite images are derived from Google Maps, with a $1,200 \times 1,200$ pixels resolution. The dataset comprises $35,532$ and $8,884$ pairs of images for training and evaluation (CVACT Val). Moreover, CVACT furnishes $92,802$ pairs of test images (CVACT Test) to buttress fine-grained urban-scale CVGL.

University-1652\cite{ref35} comprises three types of images: drone, satellite, and ground views of $1,652$ buildings from $72$ universities across the globe. On average, each building is represented by $54$ drone images, $3.38$ ground images, and one satellite image. This is the inaugural UAV-based geo-localization dataset that supports target localization and UAV navigation tasks from the UAV viewpoint.

VIGOR\cite{ref36} comprises $105,214$ ground panoramic images and $90,618$ satellite images encompassing four major United States cities (Seattle, New York, San Francisco, and Chicago). A query image of this dataset may correspond to multiple reference images, and the query image is not necessarily located at the query image center. The dataset has two particular settings: the SAME, in which images from all cities can be used in training and testing, and the CROSS, in which training is performed in Seattle and New York and evaluation is performed in Chicago and San Francisco.

CV-Cities (ours) comprises $223,736$ ground panoramic images and an equal number of satellite images all accompanied by high-precision GPS coordinates. These images represent sixteen representative cities across five continents. The ground images are $360^{\circ}$ panorama images with a resolution of $4,096 \times 2,048$ pixels, while the resolution of satellite images is $746 \times 746$ pixels, and are captured at a zoom level of $20$. The spatial resolution is $0.298 m$, corresponding to a latitude and longitude range of $0.002 \times 0.002^{\circ}$ (about $222 \times 222 m$). The images of each city in the dataset can be used for training and testing purposes.

\begin{table*}[htbp]
\caption{Comparison of CVGL datasets.}
\label{tab1}
\centering
\begin{tabularx}{\textwidth}{>{\centering\arraybackslash}p{2.4cm}
                             >{\centering\arraybackslash}p{2.1cm} 
                             >{\centering\arraybackslash}p{2.4cm} 
                             >{\centering\arraybackslash}p{2.4cm} 
                             >{\centering\arraybackslash}p{2.8cm} 
                             >{\centering\arraybackslash}p{1.8cm} 
                             >{\centering\arraybackslash}p{1.4cm} 
                             >{\centering\arraybackslash}p{0.8cm} 
                            }
\toprule
Dataset & Distribution & Training ground /sat & Test images ground /sat (Cities) & Ground-view FoV/Image size & Sat image size & GPS-tag\\
\midrule
CVUSA\cite{ref19} & USA & 35,532 / 335,532 & 8,884 / 8,884 (1) & $360^{\circ}$ / $1,232 \times 224$ & $750 \times 750$ & \checkmark \\
CVACT\cite{ref34} & Canberra & 44,416 / 44,416 & 92,802 / 92,802 (1) & $360^{\circ}$ / $1,664 \times 832$ & $1,200 \times 1,200$ & \checkmark \\
University-1652\cite{ref35} & 72 Universities & 11,663 / 701 & 5,500 / 1,652 & $512 \times 512$ & $512 \times 512$ & \texttimes \\
VIGOR\cite{ref36} & USA (4 Cities) & 51,520 / 44,055 & 53,694 / 46,563 (2) & $360^{\circ}$ / $2,048 \times 1,024$ & $640 \times 640$ & \checkmark \\
\midrule
CV-Cities (Ours) & Global (16 Cities) & 162,954 / 162,954 & 60,782 / 60,782 (6) & $360^{\circ}$ / $4,096 \times 2,048$ & $746 \times 746$ & \checkmark \\
\bottomrule
\end{tabularx}
\end{table*}

\subsection{Limitation of Current Datasets}
The most advanced CVGL methods\cite{ref8, ref37,ref38,ref39,ref40} are primarily trained and tested using CVUSA, CVACT, VIGRO, and University-1652 datasets. However, these datasets are not without limitations. For instance, although the images in the CVUSA are widely distributed in the United States, the distribution density is low, and the distance between images is considerable. While the CVACT dataset comprises more images, they are only densely distributed in Melbourne, Australia. Consequently, it is challenging for the trained models to adapt to scenarios with disparate styles globally. The distribution of images in the VIGRO is constrained to four cities within the United States, such as Chicago and New York City, which is confined within the United States. University-1652 has a limited number of images, and the images have a low resolution, lack GPS coordinates, and primarily feature buildings, which collectively prevent the model from learning universal cross-view features.

Furthermore, the test set images of CVUSA and CVACT are distributed in the same region as the training set images. This results in images in the same area having similar scene styles, such as buildings, vegetation, and roads. These similarities prevent the model from accurately demonstrating its ability in generalized CVGL. Consequently, the test accuracy is high. For instance, the top-1 accuracy of our model in CVUSA has reached $99.19\%$, rendering this dataset difficult and unchallenging as a practical test benchmark for new methods. To address the issues above, we have constructed the CV-Cities dataset, which serves as a novel training and testing benchmark for the study of CVGL.

\section{Methodology}
\subsection{CV-Cities Dataset and Overview}
\begin{figure*}[htbp]  
\centering
\begin{tabular}{c}
\includegraphics[width=0.86\textwidth]{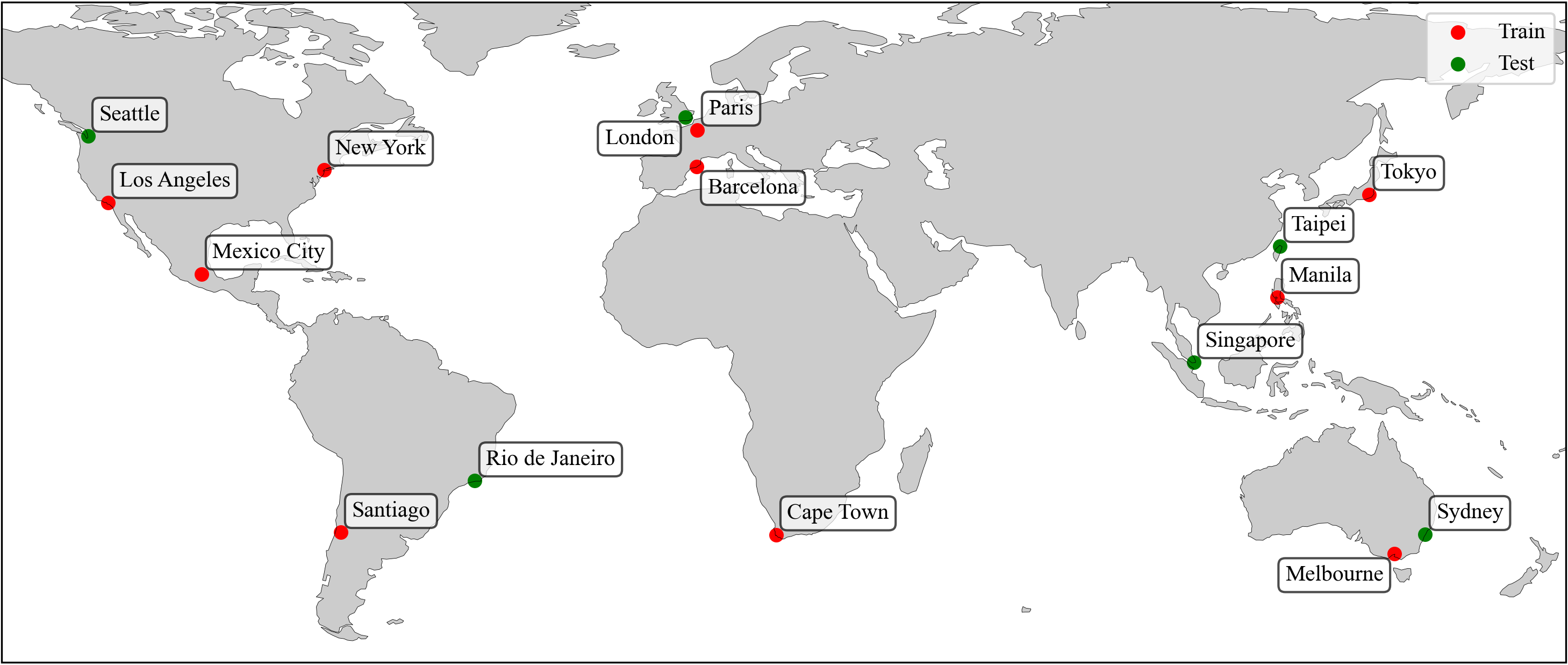}\\
(a)
\end{tabular}\\
\begin{tabular}{cccc}  
\includegraphics[width=0.20\textwidth]{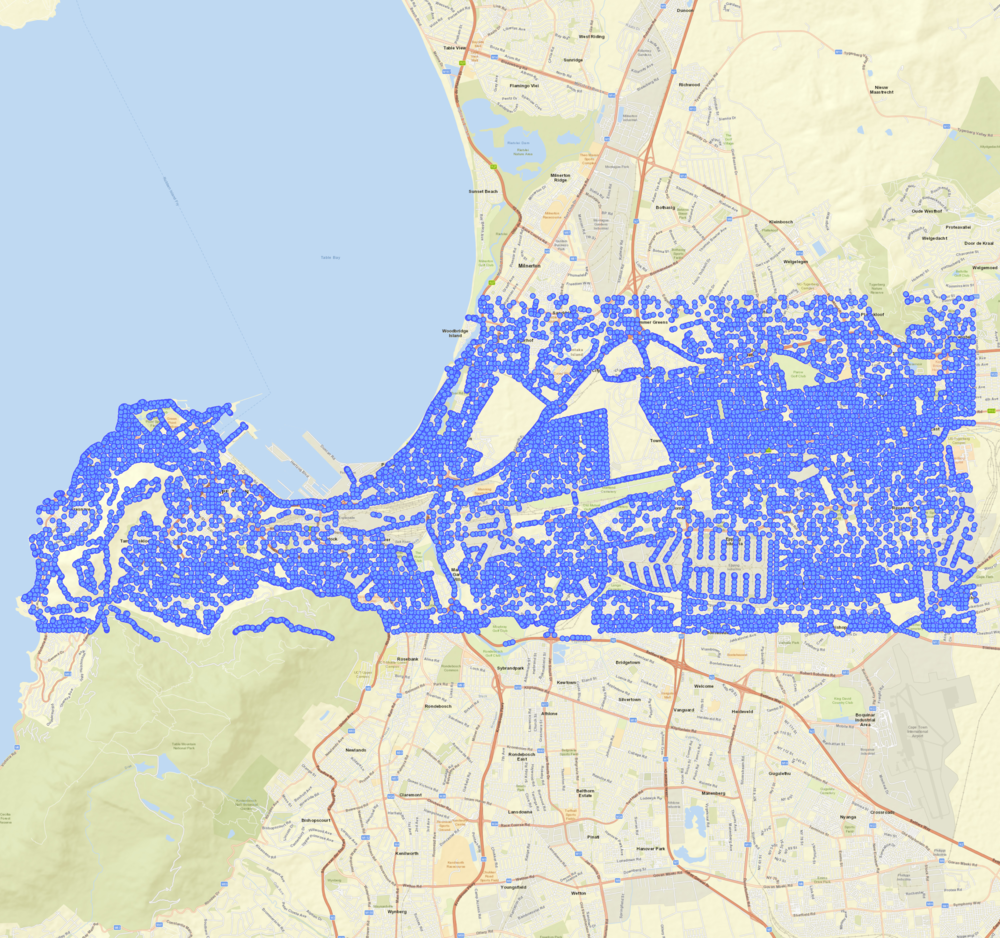} &   
\includegraphics[width=0.20\textwidth]{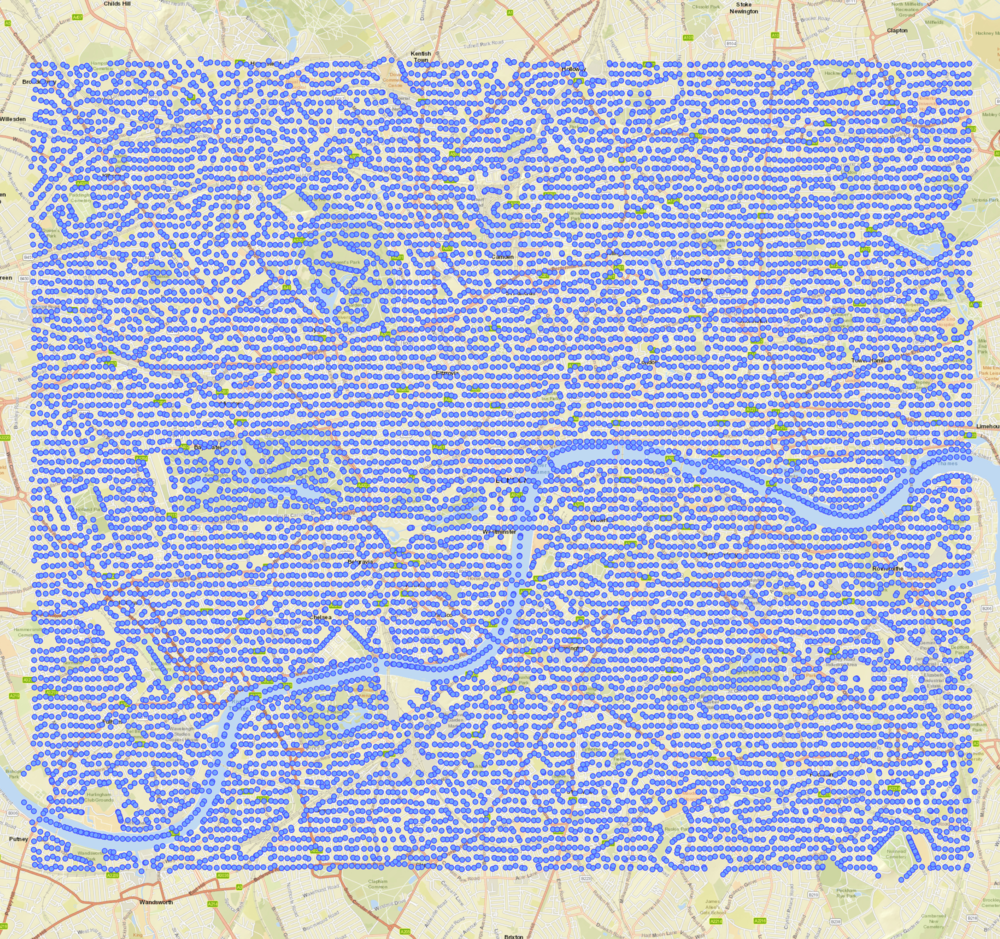} &  
\includegraphics[width=0.20\textwidth]{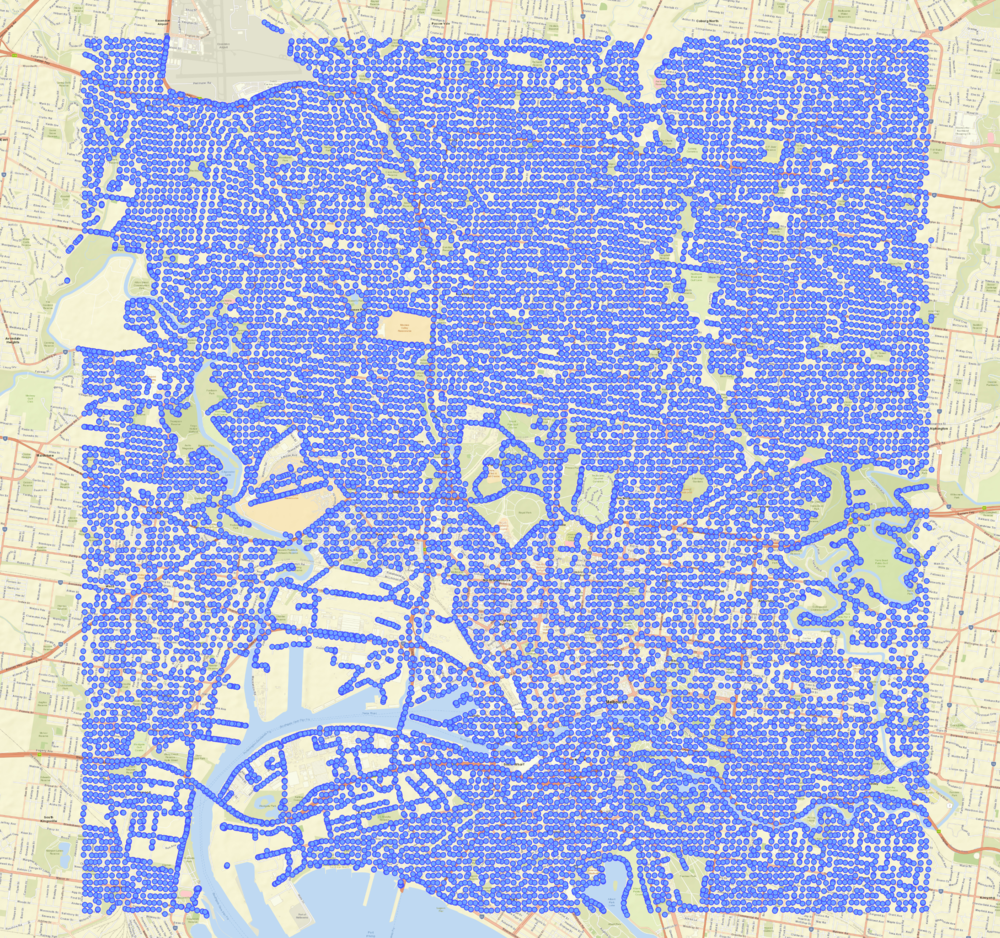} &  
\includegraphics[width=0.20\textwidth]{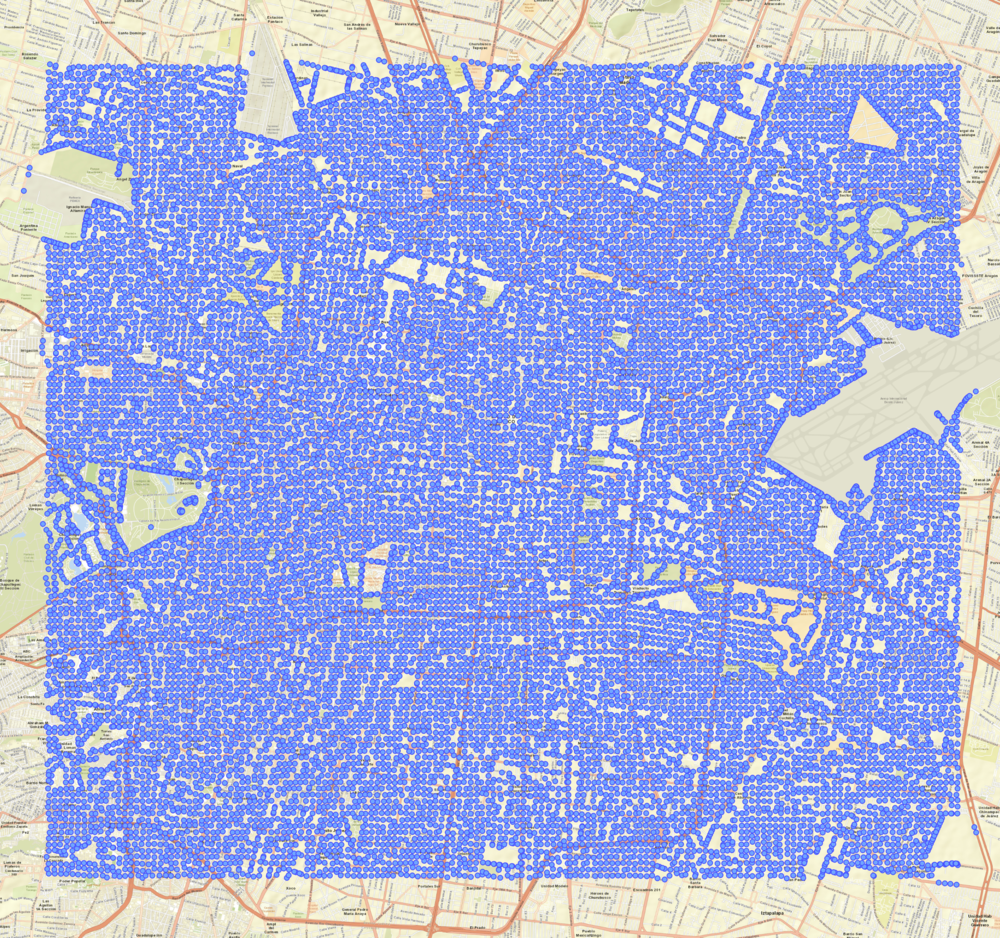} \\  
Cape Town, South Africa & London, UK & Melbourne, Australia & Mexico City, Mexico \\  
\includegraphics[width=0.20\textwidth]{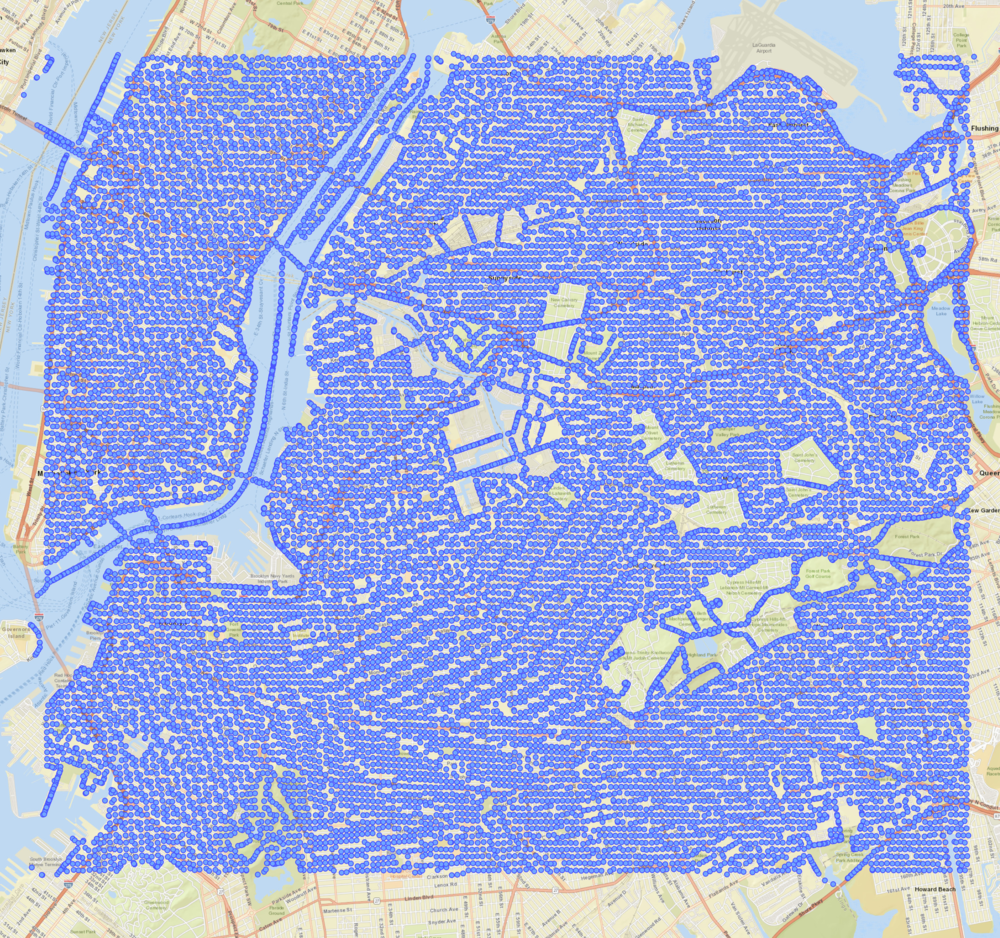} &  
\includegraphics[width=0.20\textwidth]{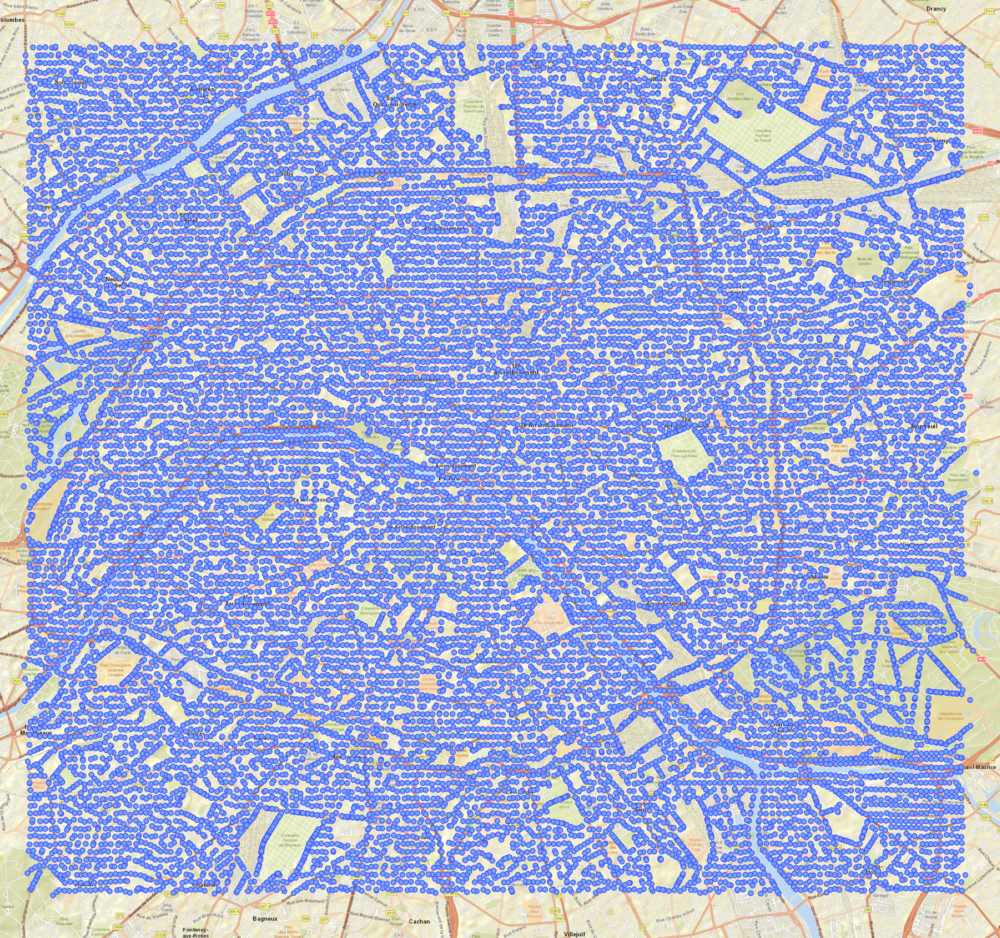} &  
\includegraphics[width=0.20\textwidth]{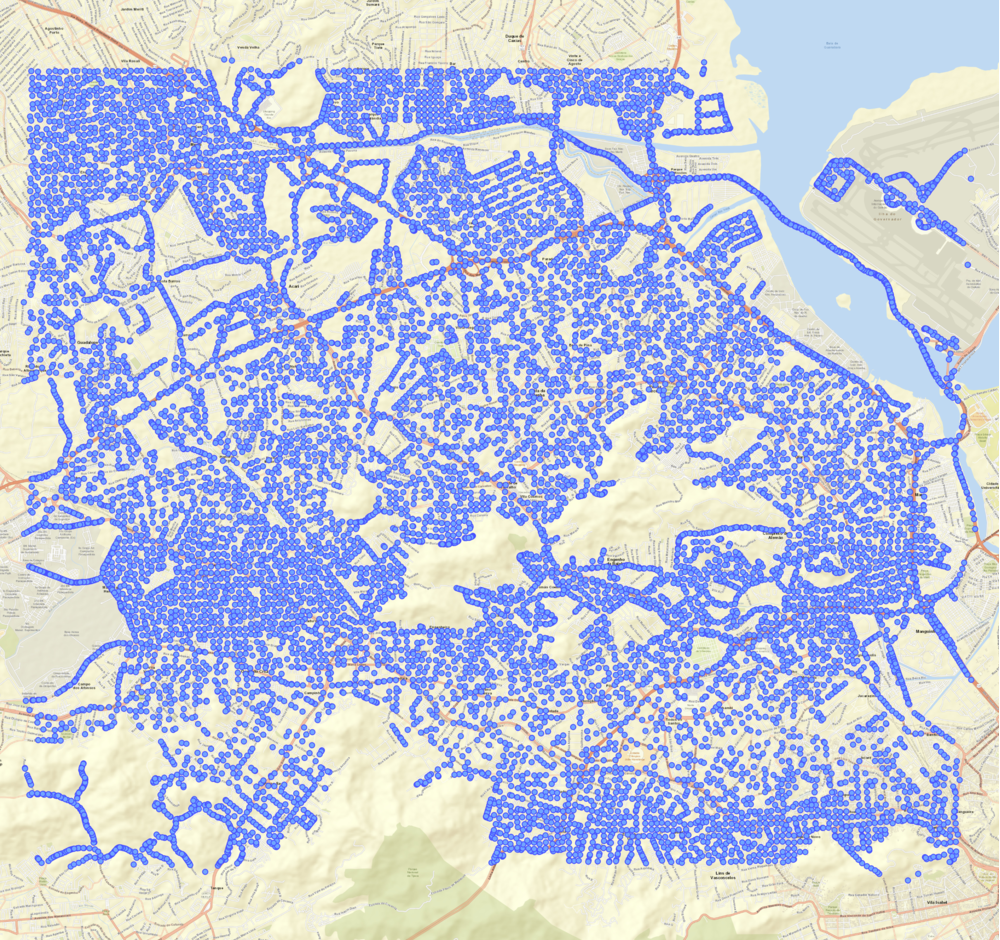} &  
\includegraphics[width=0.20\textwidth]{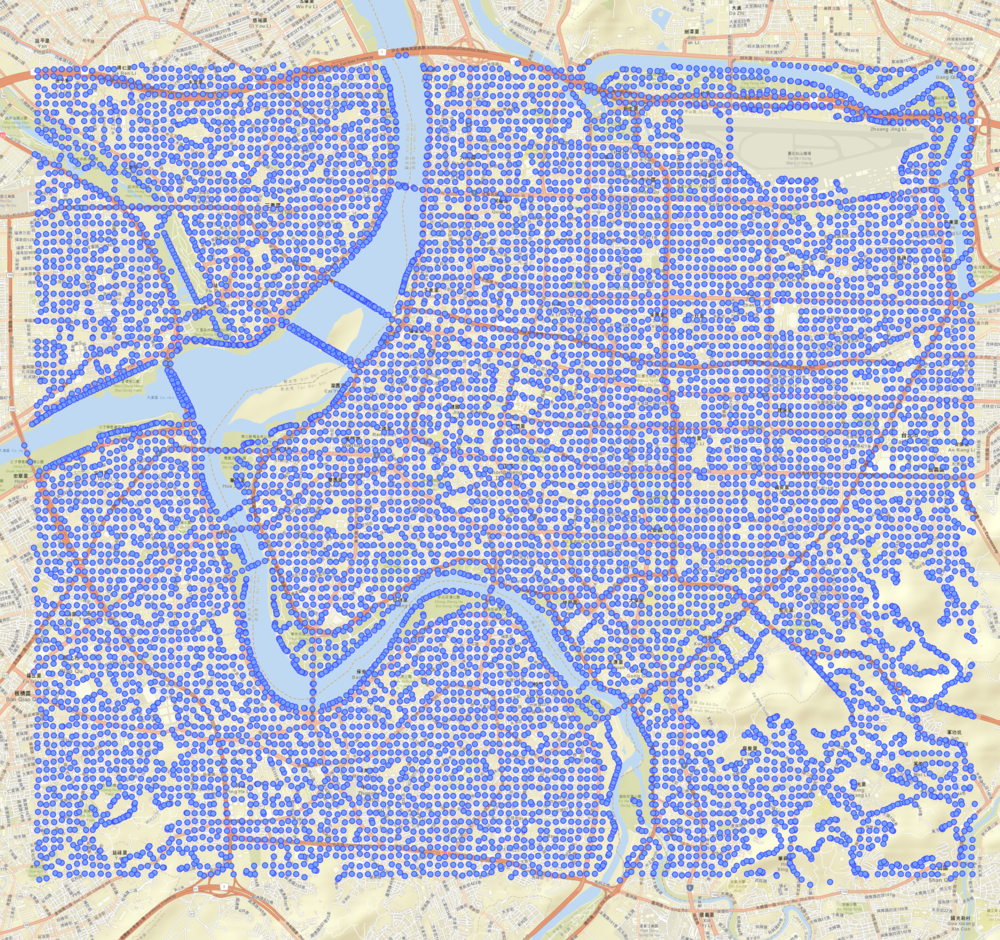} \\  
New York, USA & Paris, France & Rio de Janeiro, Brazil & Taipei, China 
\end{tabular}\\
\begin{tabular}{c}
(b)
\end{tabular} 
\caption{The city and sample points distribution map of CV-Cities datasets. (a) The city distribution map of the CV-Cities dataset, with red and green dots representing training and testing cities respectively. (b) Sample points distribution map of eight cities (out of sixteen) in the CV-Cities dataset.}
\label{fig1}  
\end{figure*}

Our CV-Cities dataset was constructed by combining ground and satellite images from Google Street View and Google Earth. To improve the geographic diversity and globalization of the data set, we selected sixteen representative cities from six continents worldwide to establish the dataset. These 16 cities cover a wide range of geographic environments from developed to developing cities, from coastal to inland, and from the southern hemisphere to the northern hemisphere. They encompass a wide spectrum of climates, including Mediterranean, highland mountainous, temperate, and tropical, each with unique vegetation and architectural characteristics. The distribution of these cities is illustrated in Fig. \ref{fig1} (a).

Ground images were collected in each city at a spacing of approximately $100$ meters. The image size was $4,096 \times 2,048$ pixels, and the field of view angle was $360^{\circ}$. The north direction of these ground images was randomized. The ground images were captured by six rolling-shutter cameras, which were either mounted on cars or carried by pedestrians. Various sensors were incorporated into the cameras to correct the image locations, ensuring that the locations of the ground images were accurate\cite{ref41}.

The satellite images are at level $20$, are $746 \times 746$ pixels in size, and have a spatial resolution of $0.298 m$. This corresponds to a latitude and longitude range of $0.002 \times 0.002^{\circ}$ (about $222 \times 222$ m). The sampling points are spaced approximately $100$ m apart, ensuring a certain degree of overlap between satellite images from neighboring sample points. The CV-Cities dataset has eight cities (sixteen in total), with the distribution of sample points in Fig. \ref{fig1} (b).

\begin{figure*}[htbp]
\centering
\begin{tabular}{
>{\centering\arraybackslash}m{0.23\textwidth}
>{\centering\arraybackslash}m{0.115\textwidth}
>{\centering\arraybackslash}m{0.23\textwidth}
>{\centering\arraybackslash}m{0.115\textwidth}
}
\includegraphics[width=\linewidth]{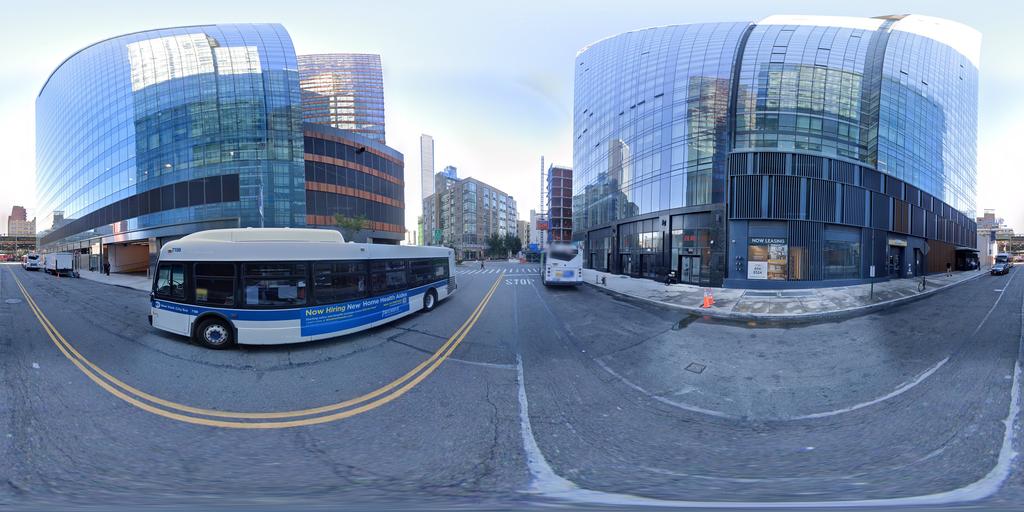} &
\includegraphics[width=\linewidth]{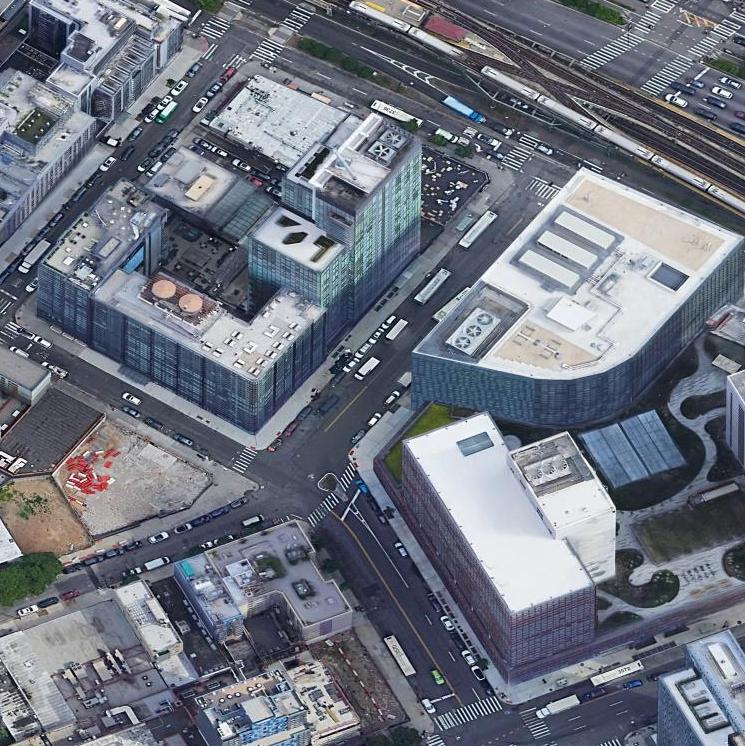} &
\includegraphics[width=\linewidth]{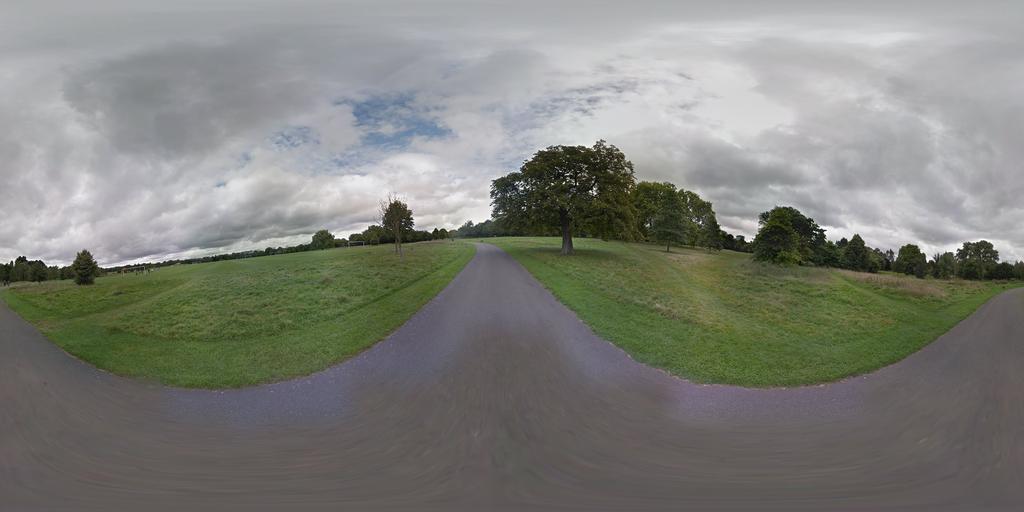} &
\includegraphics[width=\linewidth]{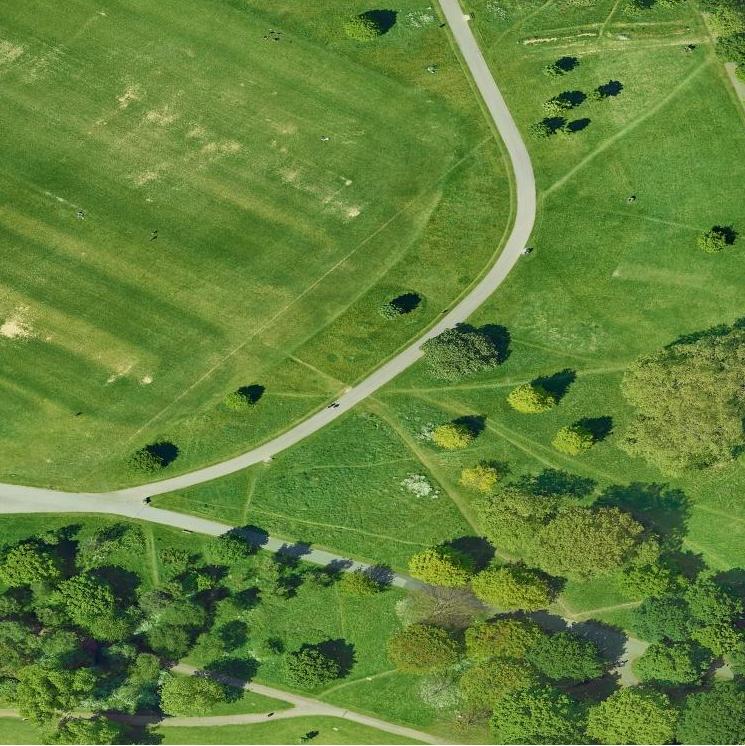} \\
\multicolumn{2}{m{0.36\textwidth}}{\centering City scene} &
\multicolumn{2}{m{0.36\textwidth}}{\centering Nature scene} \\
\includegraphics[width=\linewidth]{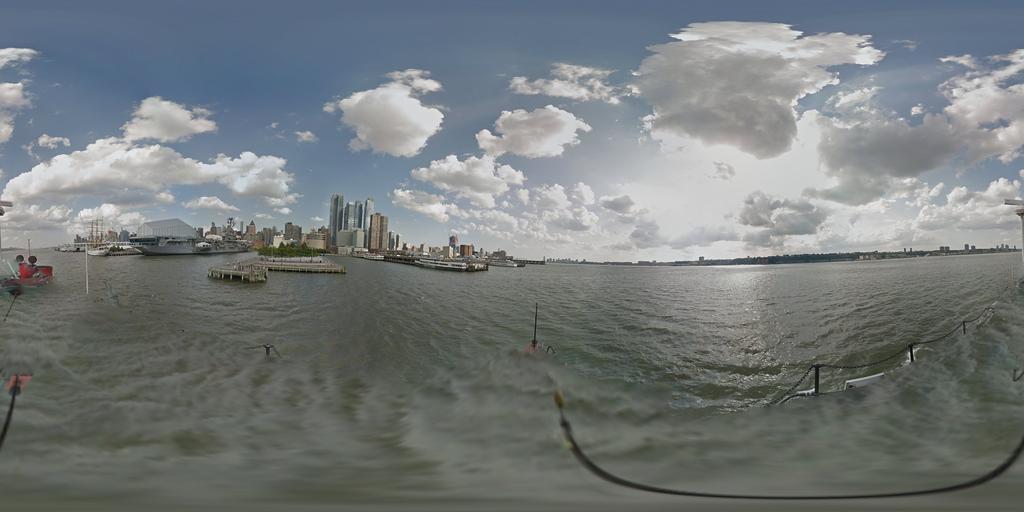} &
\includegraphics[width=\linewidth]{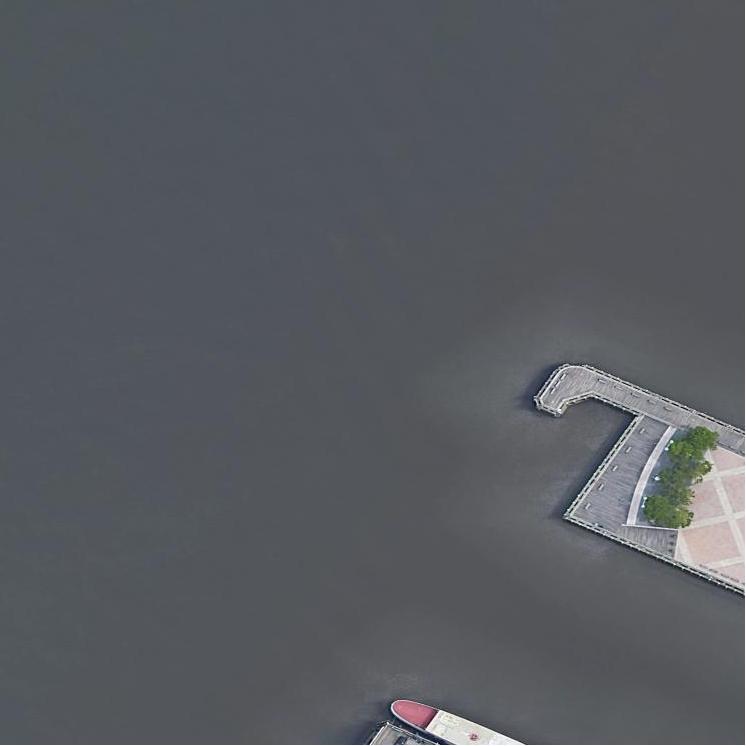} &
\includegraphics[width=\linewidth]{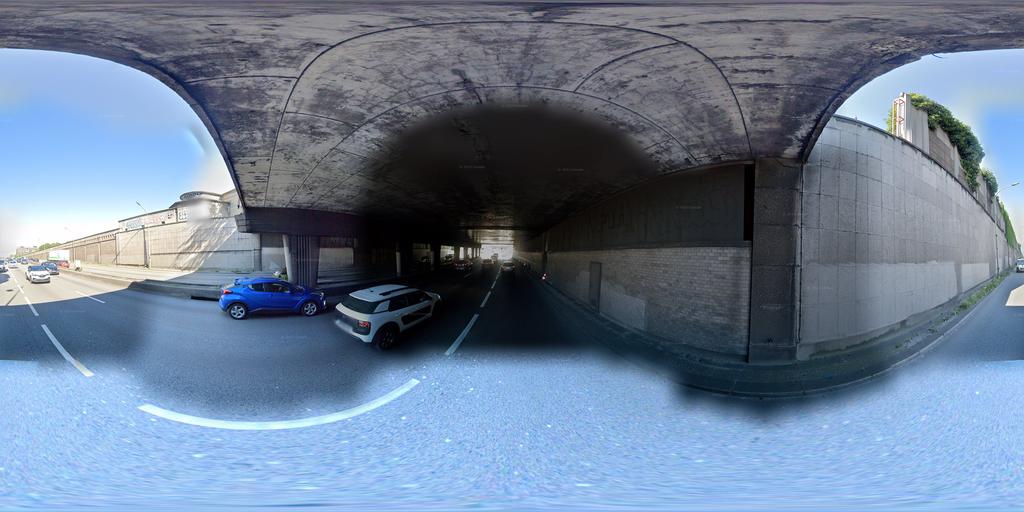} &
\includegraphics[width=\linewidth]{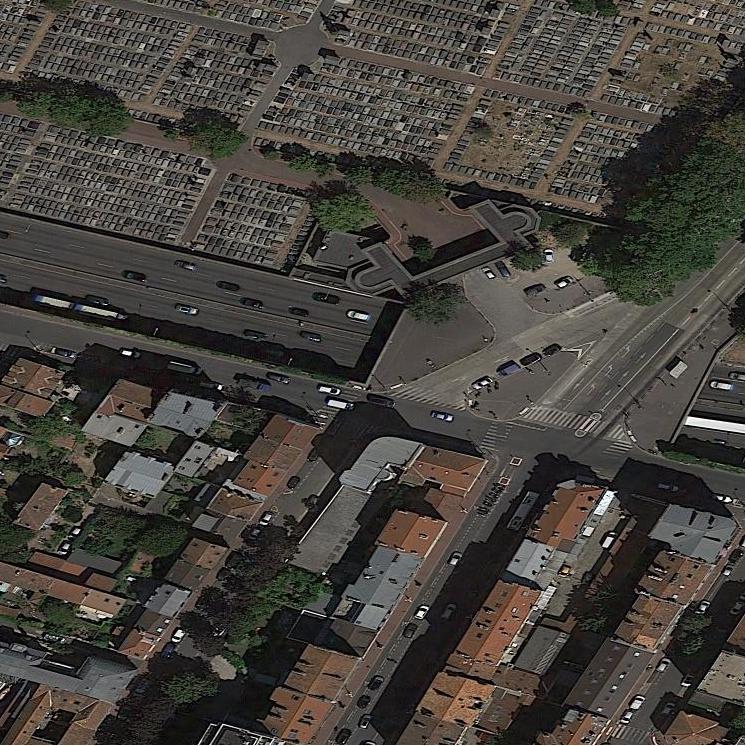} \\
\multicolumn{2}{m{0.36\textwidth}}{\centering Water scene} &
\multicolumn{2}{m{0.36\textwidth}}{\centering Occlusion} \\
\end{tabular}
\caption{Examples of ground and satellite images of different types of scenes. The left image of each scene pair is the ground image, and the right image is the corresponding satellite image.}
\label{fig2}
\end{figure*}

\begin{figure*}[htbp]
\centering
\begin{tabular}{@{}ccc@{}}
\includegraphics[width=0.3\textwidth]{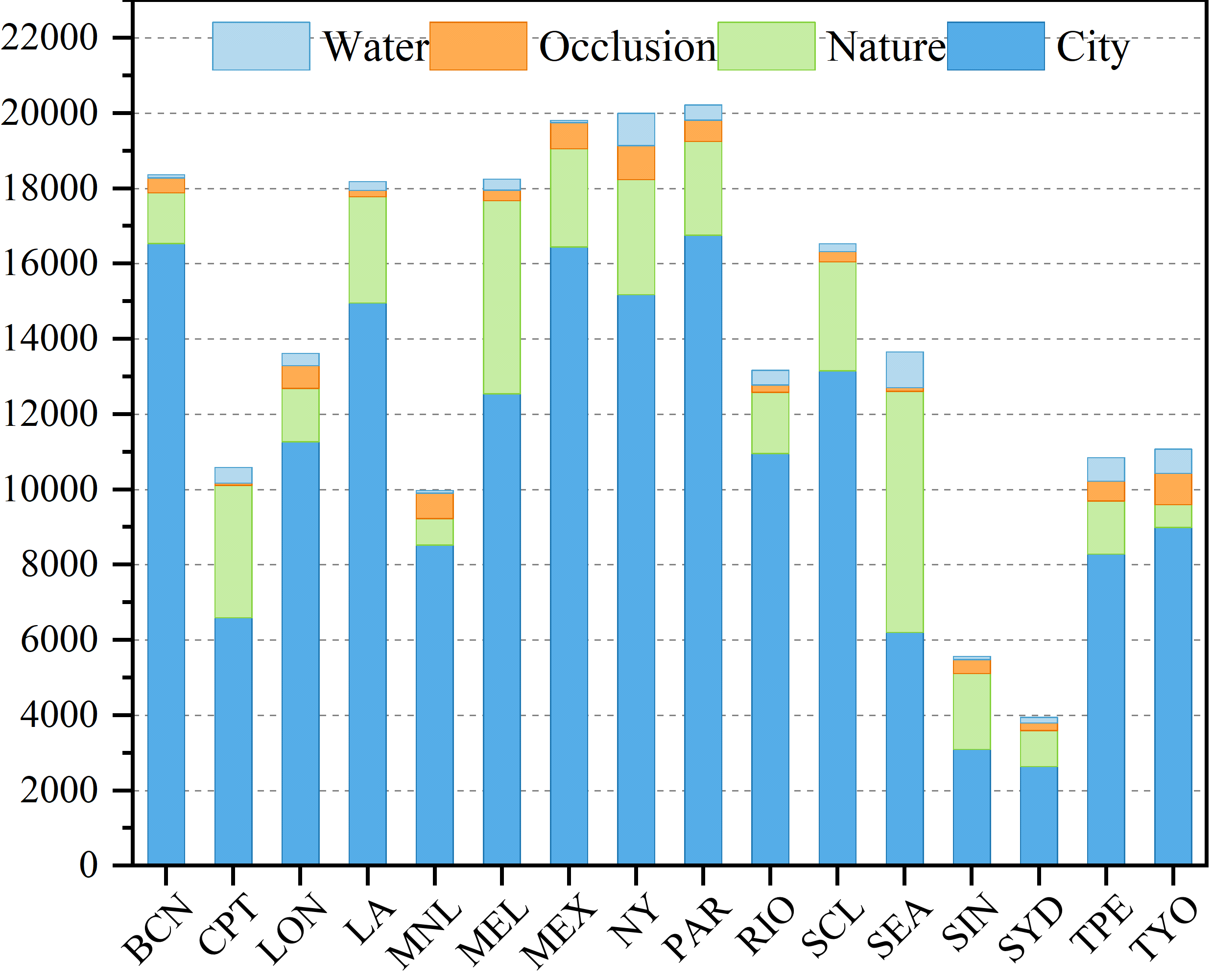} &
\includegraphics[width=0.3\textwidth]{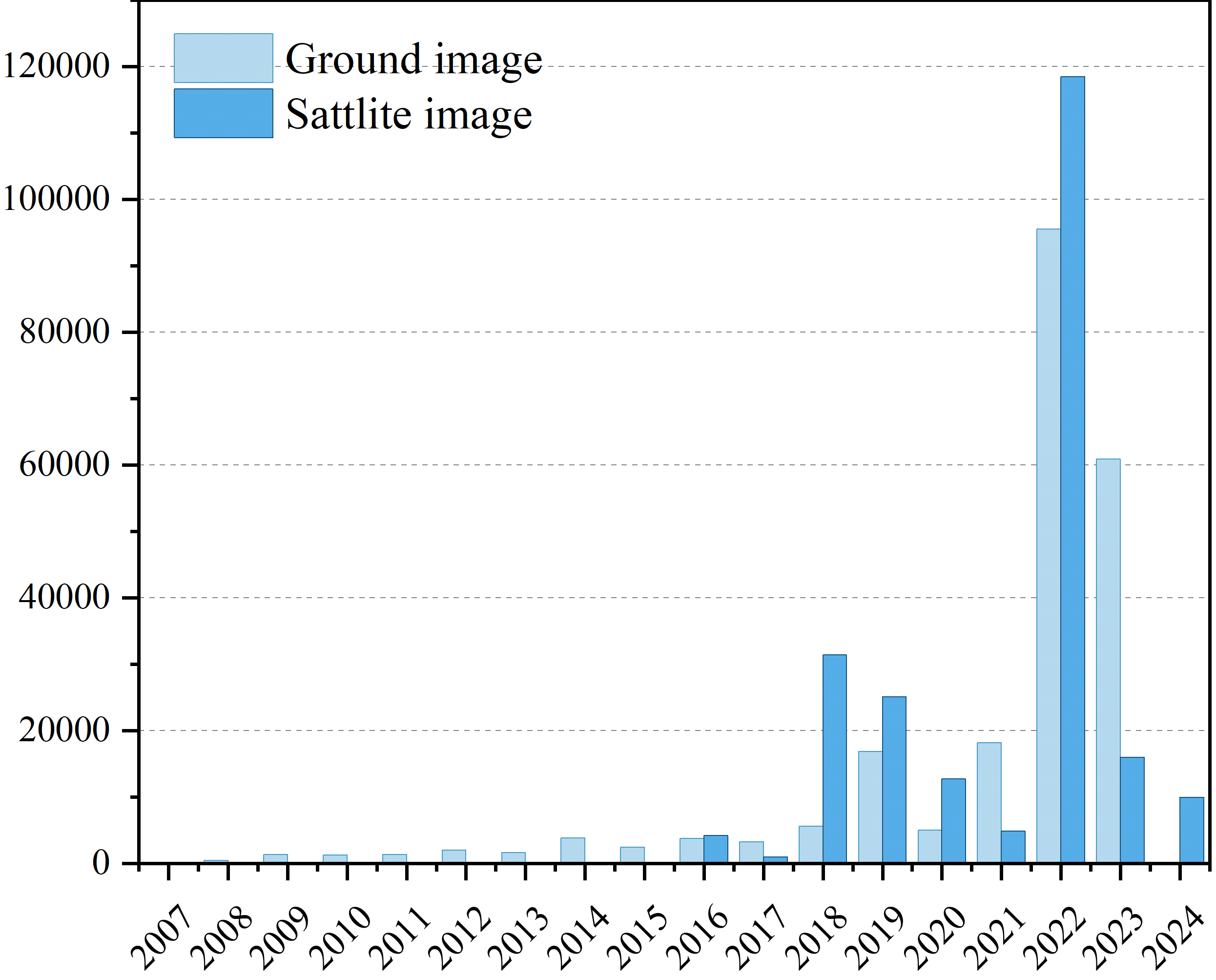} &
\raisebox{0.4em}{\includegraphics[width=0.3\textwidth]{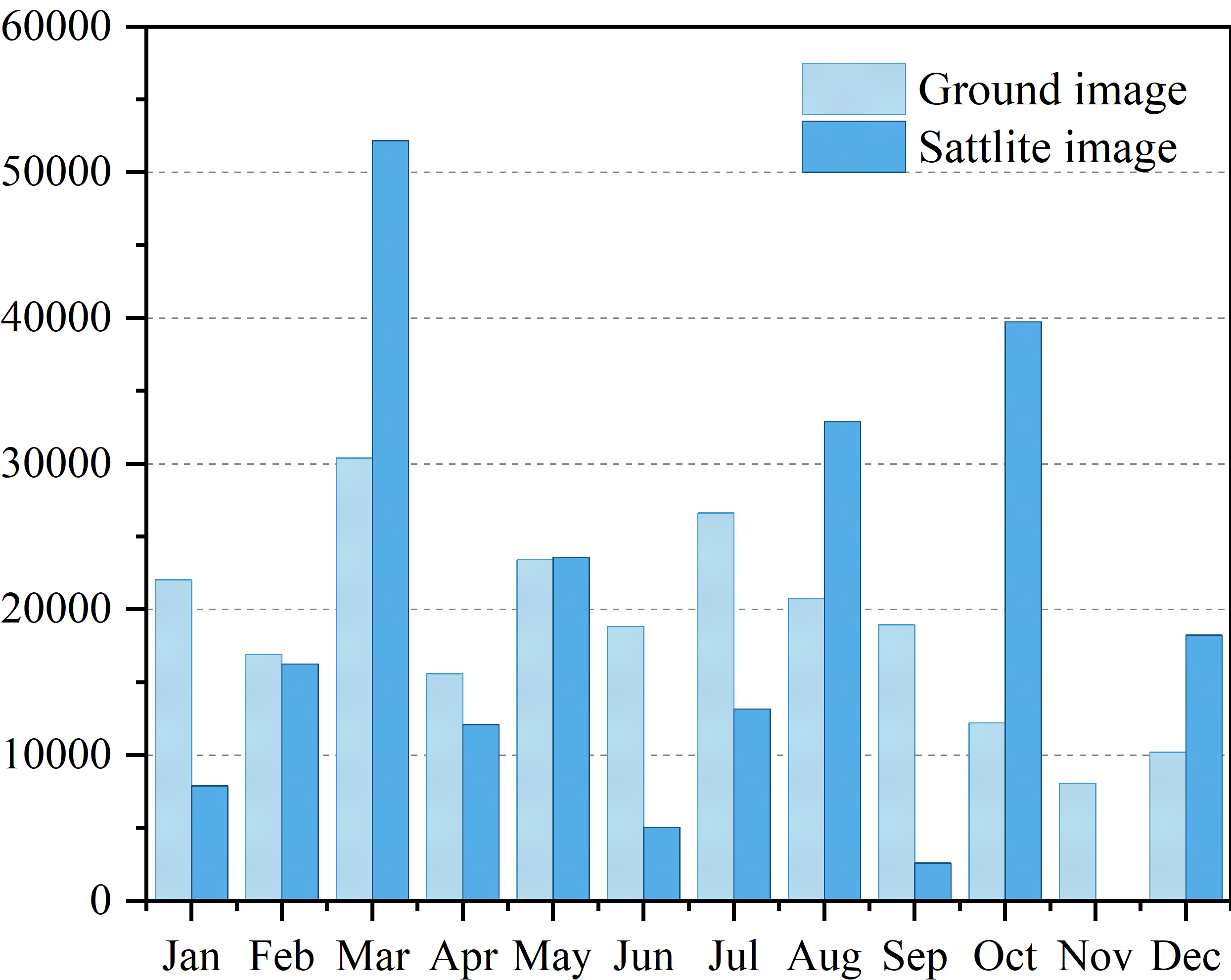}} \\
(a) & (b) & (c) \\
\end{tabular}
\caption{The distribution of images in CV-Cities on scenes, yearly and monthly scale. (a) Scenes distribution. (b) Yearly distribution. (c) Monthly distribution. }
\label{fig3}
\end{figure*}

As illustrated in Fig. \ref{fig1}, the sixteen cities of the CV-Cities are distributed across six continents of the world, with sampling points distributed evenly within each city. The sampling points encompass diverse scenes, including city streets, natural scenes, water scenes, occlusion, etc. These sample images are shown in Fig. \ref{fig2}. To obtain the scenes distribution of CV-Cities, we selected 4,000 ground images from each scene within the CV-Cities to train a ViTb16 model, aiming to accurately predict the scene categories for all sampling points in CV-Cities. As illustrated in Fig. \ref{fig3} (a), city scenes constitute the predominant category. The proportion of natural scenes is closely associated with the level of city greening and development. Additionally, the presence of water scenes is primarily influenced by the geographical location and climate type of the city. Occlusion scenes, such as those found in tunnels, subway entrances, and overpasses, are relatively scarce.

The yearly and monthly distributions of ground and satellite images of the CV-Cities were quantified. Fig. \ref{fig3} (b) illustrates the yearly distribution of ground images from $2007$ to $2023$. However, most images were acquired between $2019$ and $2023$, suggesting that the images are relatively recent. Most satellite image acquisitions occurred between $2018$ and $2024$, aligning with the timeframe of the ground images. This approach avoids the potential issue of data bias arising from the significant discrepancy in the acquisition dates of ground and satellite images. The monthly distribution of ground images in Fig. \ref{fig3} (c) demonstrates a relatively uniform pattern, with some seasonal variations. Although the acquisition months of satellite images are concentrated in March, May, August, and October, they are distributed throughout all four seasons.

In general, the data of CV-Cities is global in terms of spatial distribution, which provides an excellent foundation for the model to learn the characteristics of different geographical regions around the world. Regarding temporal distribution, the ground and satellite images were predominantly collected within the last five years. The monthly distribution is more uniform, encompassing seasonal changes, enhancing the model’s adaptability to temporal and seasonal variations. Furthermore, the scenes with rich seasonal variations make CV-Cities a new, challenging benchmark for the test of CVGL.

\subsection{Framework of CVGL}
\subsubsection{Overview of the Framework}
The framework employs the DINOv2\cite{ref6} model as the backbone for feature map extractions, addressing the challenges posed by the significant differences in viewpoints. The mix module\cite{ref7}, which is both efficient and lightweight, is then utilized to aggregate the feature maps, thereby obtaining robust global features of the images. Two sampling strategies are employed during the training process. Geographically neighboring images are selected as the initial training negative sample images using GPS coordinates. Subsequently, the visual similarity between the image features extracted by the model is leveraged to mine negative samples dynamically during the training process. The symmetric InfoNCE loss\cite{ref8, ref9, ref10} is employed in the framework to facilitate more effective model training. 

\begin{figure*}[htbp]
    \centering
    \includegraphics[width=0.7\linewidth]{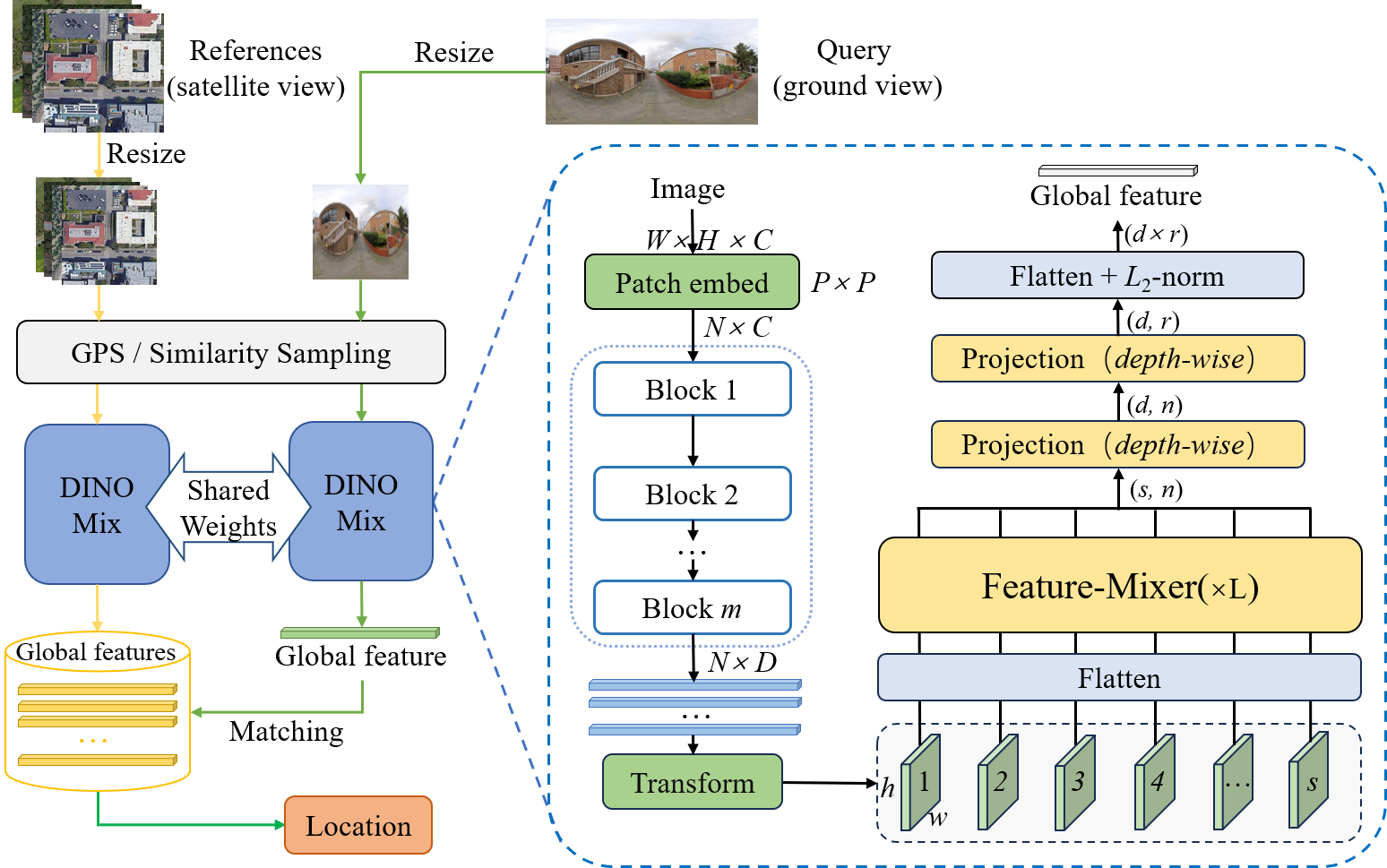}
    \caption{Our framework of CVGL.}
    \label{fig4}
\end{figure*}

Furthermore, our network weights are shared, eliminating the need to train two model weights for ground and satellite images. This results in a more lightweight network and ensures that the input sizes of images are identical, obviating the need for complex steps such as polar conversion or image generation. The CVGL framework is illustrated in Fig. \ref{fig4}.

\subsubsection{Backbone}

In CVGL tasks, the variability in image perspectives and complex scenes impose significant demands on feature extraction. To address these challenges, we use the vision foundational model DINOv2\cite{ref6} as the backbone. DINOv2 has undergone extensive training on a diverse dataset, enabling it to capture image details and features more effectively than conventional models. Its superior generalization ability ensures robust performance across various tasks, and it offers a broader range of applicable domains compared to previously released models, such as Segment Anything\cite{ref42}.

\begin{figure}[htbp]
    \centering
    \includegraphics[width=1\linewidth]{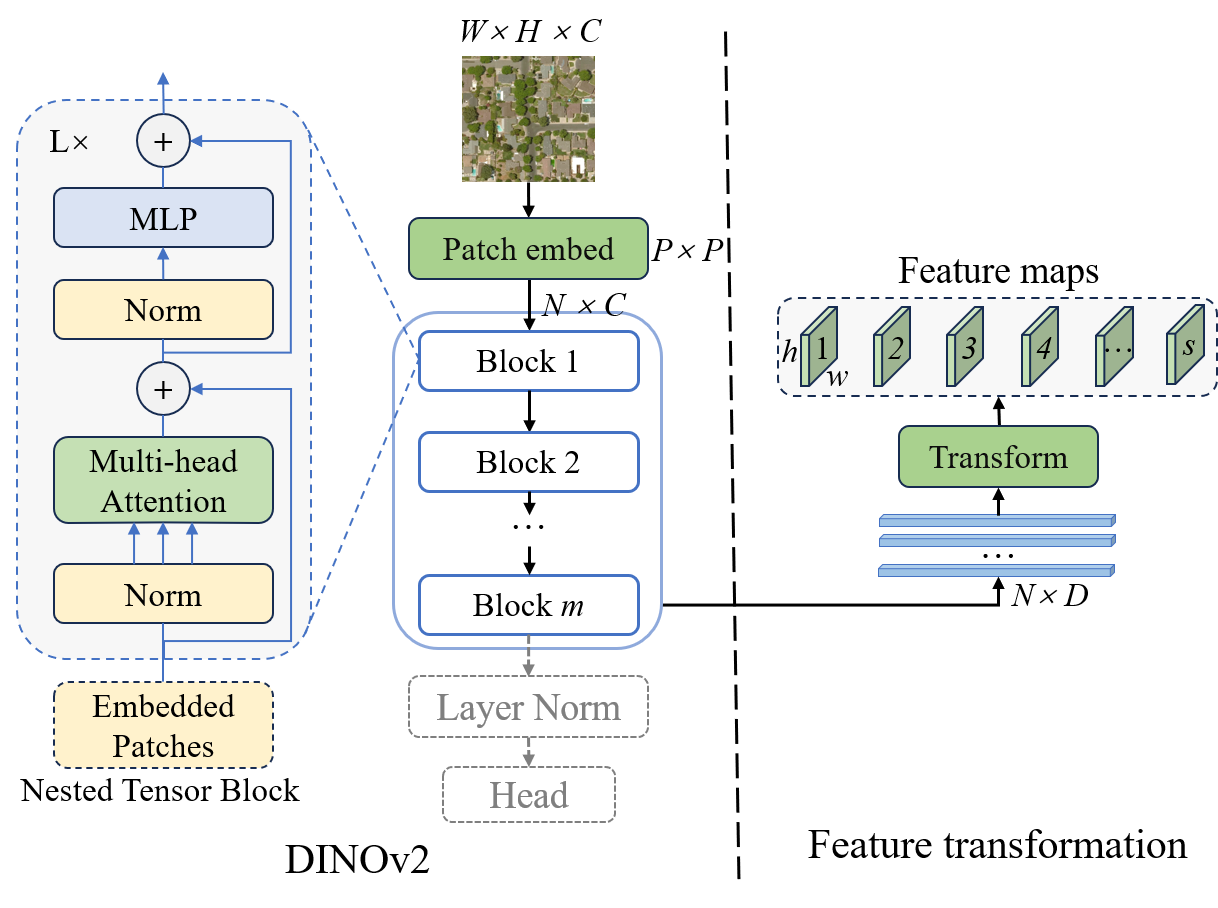}
    \caption{The structures of DINOv2 and feature transformation.}
    \label{fig5}
\end{figure}

The architecture of DINOv2 is depicted in Fig. \ref{fig5}. First, an image of size $W \times H \times C$ is input to the patch embed, which contains a $2D$ convolutional layer with a convolutional kernel of $14 \times 14$ , as well as a normal layer. This module uniformly outputs $(W // 14) \times (H // 14)$ size patches. The patches are then input to the ViT Blocks, and the number of ViT blocks varies according to the model size. Finally, the ViT Blocks output a feature matrix of size $N$ $\times$ $D$. In this study, we remove the layer normalization and head layers and add a feature transformation step to transform the $N \times D$ feature vector matrix into $S$ feature maps of size $h \times w$ as inputs to the mix module. The transformation relation is depicted in (\ref{equation1}).

\begin{equation}
\label{equation1}
    \begin{cases}
        N = hw \\
        D = s
    \end{cases}
\end{equation}

Where $N$ and $D$ are the channel count and length of the feature vector respectively, $s$, $h$, and w are the number, height, and width of the respective feature maps.

The training mode of DINOv2 is self-supervision training, which can learn from any image collection and can also learn certain features that cannot be learned by the existing methods, including a total of four sizes of ViTg14 (giant), ViTl14 (large), ViTb14 (base), and ViTs14 (small) models, among which ViTl14, ViTb14 and ViTs14 are ViTg14 obtained through knowledge distillation, inheriting the superior ability of ViTg14, with specific parameters shown in Table \ref{tab2}. After preliminary experiments, it was found that the CVGL accuracy of ViTg14 and ViTl14 as backbone was not high, and considering the limited computational resources, we primarily selected ViTb14 and ViTs14 as the backbone.

\begin{table}[htbp]
\caption{Four ViT model parameters for DINOv2.}
\label{tab2}
\centering
\begin{tabular}[c]{@{}ccccc@{}}
\toprule
Name & Size (M) & Blocks & Patch embed & Feature dim \\
\midrule
ViTg14 &  4,439.5 & 40 & 14 × 14 & 1,536  \\
ViTl14 &  1,189.0 & 24 & 14 × 14 & 1,024  \\
ViTb14 &  338.2   & 12 & 14 × 14 & 768   \\
ViTs14 &  86.2    & 12 & 14 × 14 & 384    \\
\bottomrule
\end{tabular}
\end{table}

\subsubsection{Feature Mix}
\begin{figure}
    \centering
    \includegraphics[width=0.9\linewidth]{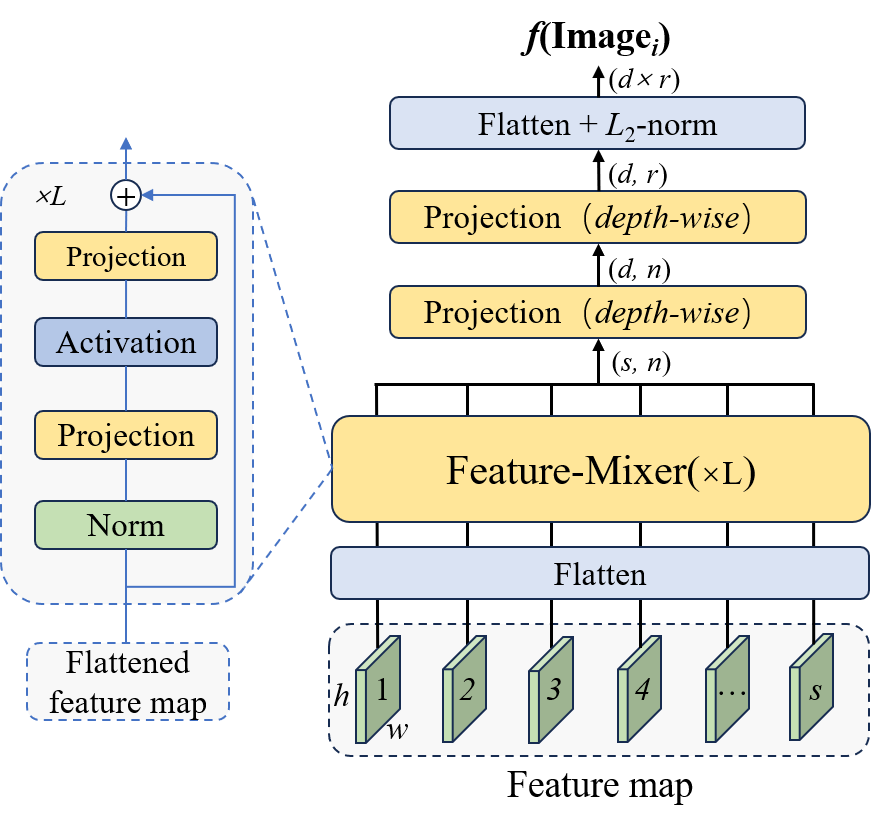}
    \caption{The architecture of the feature mix.}
    \label{fig6}
\end{figure}

Existing state-of-the-art methods primarily employ shallow aggregation layers within the feature-rich layers of cropped pre-trained backbones. However, recent advances in isotropic architectures suggest that self-attention is not essential for vision transformers. The mix module incorporates a global relationship between elements in each feature map in a cascade of feature mixing, eliminating the need for local or pyramidal aggregation. The novel method termed the ``feature mix'', offers a holistic aggregation approach that merges global relationships into each feature map through an iterative process. This technique utilizes a stack of isotropic blocks consisting of multilayer perceptrons (MLPs), collectively known as the feature mixer\cite{ref7}. This method is demonstrated to achieve high performance with minimal computational burden\cite{ref43}, as supported by various qualitative and quantitative results, as shown in Fig. \ref{fig6}.

For an input feature map $F\in R^{(s\times h\times w)}$, the mix treats the $3D$ tensor $F\in R^{(s\times h\times w)}$ as a set of $s$ $2D$ features of size h×w as shown in (\ref{equation2}):

\begin{equation}
\label{equation2}
    F=\{X^{i} \}, i=\{1, \ldots,s\}
\end{equation}

\noindent where $X^i$ is the ith feature map. Next, each $2D$ feature map $X^i$ is then flattened into a $1D$ vector, ultimately leading to an expanded feature map $F\in R^{(s\times n)}$, where $n=h\times w$.

These $s$ spread feature maps are fed into the feature mixer, a composition of $L$ MLPs with identical structures (see Fig. \ref{fig6}). The feature mixer incorporates spatial global relations into each ${X^i}\in F$ iteratively, as shown in (\ref{equation3}):

\begin{equation}
\label{equation3}
    {X^i}\gets{W_2}{\big(}\sigma({W_1}{X^i} ){\big)}+{X^i},i=\{1, \ldots,s\}
\end{equation}

\noindent where  $W_1$ and $W_2$ are the two fully connected layers’ weights that constitute the MLP, and $\sigma$ is the ReLU activation function.

Given an input matrix $F\in R^{(s\times n)}$, the feature mixer’s isotropic design generates an output matrix $Z\in R^{(s\times n)}$. This output passes through subsequent feature mixer blocks until $L$ consecutive blocks have been processed, as shown in (\ref{equation4}).

\begin{equation}
\label{equation4}
   Z={FM_L}{\Big(}FM_{(L-1)}{\big(} \ldots{FM_1} (F){\big)}{\Big)}
\end{equation}

\noindent Both $Z$ and $F$ share the same dimensions. Two fully connected layers sequentially transform the channel dimensions and rows to manage the descriptor's dimensionality. Initially, a depth projection maps $Z$ from $R^{(s\times n)}$ to $R^{(d\times n)}$ as shown in (\ref{equation5}):

\begin{equation}
\label{equation5}
    {Z^{'}}={W_d}{\big(}Transpose(Z){\big)}
\end{equation}

\noindent where $W_d$ is the fully connected layer weight. The output $Z^{'}$ from $R^{(d\times n)}$ is then mapped to $R^{(d\times r)}$ using a line-by-line projection, as shown in (\ref{equation6}):

\begin{equation}
\label{equation6}
   O={W_{r}}{\big(}Transpose({Z^{'}}){\big)}
\end{equation}

\noindent where $W_r$ is the other fully connected layer weight. The output $O$ with a dimension of $d\times r$ is subsequently expanded and $L2$ normalized to create a global feature vector.

\subsection{Sampling Strategy}
\subsubsection{Near Neighbor Sampling}
Since the model is not initially adapted to the CVGL, thus precludes the selection of negative samples that are proximal to the query images by the model. However, images with close geographic distances exhibit similarities, including architectural styles, vegetation, transportation facilities, road pavement, etc. Consequently, the near neighbor sampling strategy based on GPS location is employed during the initial training phase, enabling the selection of appropriate negative samples. The CV-Cities, CVUSA, and VIGOR images have coordinates under the WGS84 coordinate system. Consequently, the negative samples are chosen based on the haversine distance, whereas the image coordinates of CVACT are under the UTM coordinate system. In this case, the Euclidean distance is utilized to choose negative samples.

\subsubsection{Dynamic Similarity Sampling}
Following the preliminary training phase, the model has been adapted to the CVGL task. In this context, we employ Dynamic Similarity Sampling (DSS) to select negative samples for training. The initially trained model is first used to calculate the cosine distance between all samples. Subsequently, the top $S$ nearest-neighbor samples with the highest similarity are selected. These $S$ nearest-neighbor samples are then sorted by similarity, and the top $s/2$ are chosen as part of the batch. To enhance the diversity of negative samples. another $s/2$ samples are shuffled from the leftover $S-s/2$ samples Before incorporating the $s$ samples into the batch, any duplicates are removed to prevent repetition in training. The parameters are set to $s = 64$ and $S = 128$.

\subsection{Symmetric InofoNCE Loss}
An alternative to comparison learning is to utilize all available negative samples in the batch instead of the triad loss, known as InfoNCE\cite{ref44} or NTXent loss\cite{ref9}, as shown in (\ref{equation7}):

\begin{equation}
\label{equation7}
  \mathcal{L}(q, R)_{InfoNCE}=-\log \frac{exp(q\cdot r_{+}/\tau )}{\sum _{i=0}^{R}exp(q\cdot r_{i}/\tau )}
\end{equation}

In this context, $q$ is a coded ground query image, and $R$ is a set of coded satellite reference images. There is only one positive $r_i$, which is $r_+$, matching with $q$. This function computes low values for similar query-positive pairs and high values for dissimilar query-negative pairs. The loss is derived from cross-entropy and is used to gauge the similarity between views. The temperature parameter $\tau$ that can be either optimized or set to a constant value\cite{ref44}. Historically, the InfoNCE has been applied asymmetrically for unsupervised image representation learning\cite{ref9}. However, symmetric formulations have proven effective in multimodal pretraining by integrating diverse modalities\cite{ref44}. Consequently, this loss function is now applied symmetrically to facilitate bidirectional information flow between satellite and ground images. Each positive sample is contrasted with $N-1$ negative samples ($N$ is batch size). This allows for the simultaneous definition of multiple instances.

\section{Experiments}
This section presents the specifics of the training and evaluation processes. After that, a comparison is made between our and state-of-the-art methodology. Subsequently, ablation studies are conducted to assess the impact of the backbone structure, the number of mix layers, and the model’s generalization capabilities. Subsequently, the CVGL results for several images are presented, and the precision distribution of our framework in each city is illustrated and analyzed. Finally, heat maps of sample images of typical scenes are generated to analyze the framework’s performance qualitatively.

\subsection{Datasets and Metrics}
Unless otherwise stated, we use ten city images from CV-Cities for training by default and another six cities to test the model’s CVGL accuracy. Our framework is compared with state-of-the-art on five benchmarks: CV-Cities, CVUSA, CVACT, Universities-1652, and VIGOR. University-1652 contains two modes, Drone2Sat and Sat2Drone, and VIGOR comprises two modes, SAME and CROSS.

To compare the accuracy of our method with previous approaches, we employed three evaluation metrics: top-k, average precision (AP), and hit rate, across various datasets. Top-k \cite{ref33, ref45, ref46, ref47} is the metric used to calculate the percentage of predicted locations that the top-k locations with the highest probability contain correct locations. Specifically, when a list of candidate locations is generated for a query image using the geo-localization method, the image is considered correctly localized if one of the top-k candidates matches the ground truth. Top-k is more suitable for datasets with one-to-one ground-truth matching, as expressed in (\ref{equation8}):

\begin{equation}
\label{equation8}
  TopK = \sum _{i=1}^{n}S_{i, k}/n
\end{equation}
Where $n$ is the number of query images, $i$ is the query image labeled as $i$. If there is a correct image in the first $k$ results, $S_{(i,k)}$ is $1$, otherwise it is $0$. The commonly used ones are top1, top5, top10, and top1\%.

However, in the University-1652, one satellite image corresponds to multiple ground images. Top-k does not adequately reflect the matching results of the rest ground-truth images. Therefore, AP is a better choice. AP\cite{ref18, ref35} is an integral of the precision of each recall threshold, that is, the sum of the areas under the PR curve, as shown in (\ref{equation9}):

\begin{equation}
\label{equation9}
  AP = \sum _{i=1}^{n} (R_i-R_{i-1}) P_i
\end{equation}
Where $n$ is the number of query images, $R_i$ is the first recall threshold, $P_i$ is the precision rate.

Hit rate\cite{ref39}, if the retrieved top1 reference image overrides the query image (including ground truth), it is considered a “hit”, the percentage of hits to the total number of queries is the hit rate.

\subsection{Implementation Details}
\subsubsection{Model settings}
Given the many parameters and training efficiency, we have elected to utilize the ViTb14 and ViTs14 of DINOv2 as the backbone. To this end, we have removed the final layer of normalization and the head layer, incorporated a feature transformation layer to generate the feature maps, and evaluated the framework’s performance with ViTb14 and ViTs14 as a backbone. To meet the experimental design requirements, we gradually release the parameter updates from the mix module to ViT blocks and then to the patch embedding module, aiming to identify the optimal model parameter updating strategy.

\subsubsection{Training}
To complete our experiments, we utilized six RTX4090 GPUs. During model training, we incorporated a label smoothing setting of $0.1$ within the InfoNCE loss to prevent the model from overfitting to the training dataset. Since the north direction of the ground query image is not defined during the geo-localization process, a rotation of the north direction of the ground panorama image is performed using horizontal cropping and splicing with a probability of $75\%$ during the training process. This is coupled with a simultaneous rotation of the satellite image to align more closely with the intended application. Random image compression, color jitter, blurring, and sharpening are applied to the ground and satellite images. Random grid-based pixel masking also simulates image occlusion, enhancing the model’s generalizability. The model was trained using the SGD optimizer with a batch size of $32$. The initial learning rate was set to $0.001$, employing a cosine learning rate scheduler with a 1-epoch warm-up period. The training process spanned $40$ epochs uniformly.

\subsubsection{Evaluation}
Our evaluation metrics are consistent with existing papers\cite{ref33, ref45, ref46, ref47} Top-k is the evaluation metric for all datasets in this paper. For University-1652, we use the AP\cite{ref18, ref35} to evaluate the model precision. We also use Hit Rate\cite{ref39} as an evaluation metric for VIGOR. It should be noted that all numbers in \textbf{bold} and \underline{underlined} in the tables of this paper represent optimal and suboptimal results respectively.

\subsection{Comparison to the State-of-the-art}
\label{subection_Comparison_to_the_State-of-the-ar}
In this section, we compare our framework with state-of-the-art on CV-Cities, CVUSA, CVACT, University-1652, and VIGOR, respectively, to demonstrate the advantages of our framework in CVGL, as shown in Table \ref{tab3}, Table \ref{tab4} and Table \ref{tab5}.

\begin{table*}[htbp]
\centering
\caption{Comparison with the state-of-the-art results on CV-Cities, CVUSA, and CVACT. Shared W is whether the weights are shared or not, \#Para is the parametric size of the model, and CV-Cities avg. is the average test accurates of the six cities.}
\label{tab3}
\begin{tabular}{ccccccccccc}
\toprule
\multirow{2}{*}{Methods} & \multicolumn{1}{c}{\multirow{2}{*}{Shared W}} & \multicolumn{1}{c}{\multirow{2}{*}{\#Para(M)}} & \multicolumn{2}{c}{CV-Cities avg.} & \multicolumn{2}{c}{CVUSA} & \multicolumn{2}{c}{CVACT Val} & \multicolumn{2}{c}{CVACT Test} \\ \cline{4-11} \addlinespace
& \multicolumn{1}{c}{} & \multicolumn{1}{c}{} & \multicolumn{1}{c}{Top1(\%)} & \multicolumn{1}{c}{Top5(\%)} & \multicolumn{1}{c}{Top1(\%)} & \multicolumn{1}{c}{Top5($\%$)} & \multicolumn{1}{c}{Top1($\%$)} & \multicolumn{1}{c}{Top5($\%$)} & \multicolumn{1}{c}{Top1($\%$)} & \multicolumn{1}{c}{Top5($\%$)} \\ 
\midrule
LPN\cite{ref38} & $\times$ & $138.7\times 2$ & - & - & 85.79 & 95.38 & 79.99 & 90.63 & - & - \\
SAFA\cite{ref2} & $\times$ & $14.7\times 2$ & 22.42 & 41.82 & 89.84 & 96.93 & 81.03 & 92.80 & - & - \\ 
TransGeo\cite{ref39} & $\times$ & $22.4\times 2$ & 26.19 & 52.29 & 94.08 & 98.36 & 84.95 & 94.14 & - & - \\ 
GeoDTR\cite{ref28} & $\times$ & $21.3\times 2$ & 32.23 & 55.41 & 93.76 & 98.47 & 85.43 & 94.81 & 62.96 & 87.35 \\ 
SAIG-D\cite{ref40} & $\times$ & $15.6\times 2$ & 42.21 & 68.73 & 96.08 & 98.72 & 89.21 & 96.07 & - & - \\ 
Sample4Geo\cite{ref8} & $\checkmark$ & 88.6 & 74.49 & 84.07 & 98.68 & 99.68 & \underline{90.81} & 96.74 & \underline{71.51} & \underline{92.42} \\ \midrule
ViTb14-mix(ours) & $\checkmark$ & 90.5 & \textbf{82.91} & \textbf{90.14} & \textbf{99.19} & \textbf{99.80} & \textbf{92.59} & \textbf{97.16} & \textbf{73.70} & \textbf{94.28} \\
ViTs14-mix(ours) & $\checkmark$ & 22.7 & \underline{75.01} & \underline{85.13} & \underline{98.69} & \underline{99.73} & 90.71 &\underline{96.90} & 68.16 & 92.42 \\
\bottomrule
\end{tabular}
\end{table*}

From Table \ref{tab3}, the top-1 accuracies of our ViTb14-mix in CVUSA, CVACT Val, and CVACT Test exceed the current state-of-the-art method, reaching $82.91\%$, $99.19\%$, $92.59\%$, and $73.70\%$, respectively. It is evident that the test accuracies of CVUSA and CVACT Val, which are the two benchmarks, are saturated with accuracy and cannot effectively reflect the capability of the new CVGL framework. Although the ViTb14-mix model has a slightly larger parameter count ($90.5 M$) compared to Sample4Geo, its significant accuracy improvements justify the increase. The lightweight ViTs14-mix, with only $22.7 M$ parameters, also shows competitive performance, offering a balanced trade-off between model complexity and accuracy.

\begin{table}[htbp]
\centering
\caption{Comparison results on University-1652.}
\label{tab4}
\begin{tabular}{ccccc}
\toprule
\multirow{2}{*}{Methods} & \multicolumn{2}{c}{Drone2Sat}& \multicolumn{2}{c}{Sat2Drone} \\ \cline{2-5} \addlinespace
 & Top1($\%$) & Ap ($\%$) &Top1($\%$) & Ap ($\%$) \\ 
\midrule
LPN\cite{ref38} & 75.93 & 79.14 & 86.45 & 74.79 \\
SAIG-D\cite{ref40} & 78.85 & 81.62 & 86.45 & 78.48 \\ 
MBF\cite{ref48} & 89.05 & 90.61 & 92.15 & 84.45\\ 
Sample4Geo\cite{ref8} & 92.65 & 93.81 & 95.14 & 91.39 \\ \midrule
ViTb14-mix(ours) & \textbf{97.43} & \textbf{95.01} & \underline{96.01} & \underline{92.57} \\
ViTs14-mix(ours) & \underline{93.76} & \underline{94.78} & \textbf{96.15} & \textbf{92.94} \\
\bottomrule
\end{tabular}
\end{table}

Table \ref{tab4} illustrates the performance of our model on the University-1652. In the Drone2Sat task, the top-1 accuracy and AP of the ViTb14-mix are enhanced by $4.78\%$ and $1.2\%$, respectively, compared to the preceding state-of-the-art, Sample4Geo. Additionally, the ViTs14-mix also exhibits a notable performance in this task. In the Sat2Drone mission, the ViTb14-mix and ViTs14-mix once more exhibit their exceptional performance. Even though Sample4Geo also performs well in the Sat2Drone task, it is still not as advanced as our ViTb14-mix and ViTs14-mix. This demonstrates that our framework not only in CVGL tasks between ground and satellite views but also shows strong performance in CVGL tasks between drone and satellite views.

\begin{table}[htbp]
\centering
\caption{Comparison results on VIGOR.}
\label{tab5}
\begin{tabular}{ccccc}
\toprule
 Mode & Methods & Top1($\%$) & Top5($\%$) & Hit Rate($\%$) \\
\midrule
\multirow{6}{*}{SAME} & LPN\cite{ref38} & 33.93 & 58.42 & 36.87 \\
& SAIG-D\cite{ref40} & 61.48 & 87.54 & 73.09 \\ 
& MBF\cite{ref48} & 65.23 & 88.08 & 74.11 \\ 
& Sample4Geo\cite{ref8} & \underline{77.86} & \underline{95.66} & \underline{89.92}  \\ \cline{2-5} \addlinespace
& ViTb14-mix(ours) & \textbf{78.27} & \textbf{96.10} & \textbf{90.76}  \\
& ViTs14-mix(ours) & 72.04 & 92.35 & 82.50 \\ \midrule
\multirow{6}{*}{CROSS} & LPN\cite{ref38} & 8.20 & 19.59 & 8.85 \\
& SAIG-D\cite{ref40} & 18.99 & 38.24 & 21.21 \\ 
& MBF\cite{ref48} & 33.05 & 55.94 & 36.71 \\ 
& Sample4Geo\cite{ref8} & \underline{61.70} & \underline{83.50} & \underline{69.87}  \\ \cline{2-5} \addlinespace
& ViTb14-mix(ours) & \textbf{64.61} & \textbf{87.48} & \textbf{75.97}  \\
& ViTs14-mix(ours) & 58.82 & 82.84 & 68.50 \\
\bottomrule
\end{tabular}
\end{table}

Table \ref{tab5} shows the performance of our model on VIGOR. The ViTb14-mix outperforms the state-of-the-art in all evaluation metrics, including top1 and top5 accuracies and hit rates in both SAME and CROSS modes. The significant improvement in CROSS mode, where the training and testing cities are separated, demonstrates our model's capability to better learn the generalizable relationships between ground and satellite images. This improvement can be attributed to our framework's ability to capture complex scene-level information and align features more effectively across disparate views, leading to a $3.84\%$ improvement over Sample4Geo in hit rate for the CROSS setting. The ViTs14-mix also surpasses existing methods with fewer parameters, further reinforcing our model's competitiveness in CVGL tasks.

The enhanced performance across all datasets stems from the synergy between our architectural design, sampling strategy, and loss function. The ViTb14-mix and ViTs14-mix  leverage the pretrained foundational vision backbone and the mix module to effectively aggregate features across views. The combined NNS and DSS strategies further boost training robustness by selecting high-quality negative samples. Additionally, symmetric InfoNCE loss improves feature alignment, especially in datasets with large intra-class variations like CV-Cities and VIGOR, highlighting the robustness of our framework in diverse cross-view scenarios.

\subsection{Ablation studies}
\subsubsection{Backbone Architecture}
The underlying backbone significantly influences the accuracy of CVGL. We trained and tested the model with ViTb14 and ViTs14 as the backbone, with the number of feature mixer layers set to two, respectively, on the CV-Cities, and the results are shown in Table \ref{tab6}. The test results demonstrate that the model with ViTb14 as the backbone exhibits superior accuracy compared to ViTs14 across all six test cities.

\begin{table}[htbp]
\centering
\caption{The test accuracy of ViTb14-mix and ViTs14-mix on the CV-Cities.}
\label{tab6}
\begin{tabular}{ccccc}
\toprule
\multirow{2}{*}{City} & \multicolumn{2}{c}{ViTb14-mix}& \multicolumn{2}{c}{ViTs14-mix} \\ \cline{2-5} \addlinespace
 & Top1($\%$) & Top5($\%$) &Top1($\%$) & Top5($\%$) \\ 
\midrule
Taipei & 60.10 & 75.19 & 47.17 & 63.56 \\
London & 91.80 & 96.00 & 81.41 & 90.96 \\
Rio & 88.34 & 95.11 & 81.20 & 90.97 \\
Seattle & 87.73 & 92.33 & 81.88 & 89.39 \\
Sydney & 92.21 & 96.22 & 88.91 & 94.59 \\
Singapore & 77.21 & 85.98 & 69.45 & 81.33 \\
Avg. & 82.91 & 90.14 & 75.00 & 85.13 \\
\bottomrule
\end{tabular}
\end{table}

Following the identification of ViTb14 as the principal component of the model, the model was trained and tested with seven distinct parameter update combinations. The combinations above are presented in Table \ref{tab7}. The seven combinations are as follows: 1. mix denotes updating only the parameters of the mix module. 2. Last n blocks + mix denotes updating the parameters of the last $n$ $(n = 2, 4, 6, 8)$ blocks and mix modules of the backbone. 3. All blocks + mix denotes updating the parameters of all twelve blocks and mix modules of the backbone. 4. All denotes updating all the parameters of the model. 

\begin{table*}[htbp]
\centering
\caption{Results of ablation studies with different combinations of parameter updates for the ViTb14-mix.}
\label{tab7}
\begin{tabular}{ccccccc}
\toprule
\multirow{2}{*}{Training blocks} & \multicolumn{2}{c}{CV-Cities Avg.}& \multicolumn{2}{c}{CVUSA} & \multicolumn{2}{c}{University-1652(D2S)} \\ \cline{2-7} \addlinespace
 & Top1($\%$) & Top5($\%$) &Top1($\%$) & Top5($\%$) &Top1($\%$) & AP($\%$) \\ 
\midrule
Mix & 8.54 & 20.27 & 64.01 & 84.94 & 85.78 & 88.15\\
Last 2 blocks + mix & 50.63 & 69.82 & 98.66 & 99.70 & 93.70 & 94.84\\
Last 4 blocks + mix & 62.19 & 78.26 & 98.87 & 99.73 & \textbf{97.43} & 95.01\\
Last 6 blocks + mix & 74.66 & 85.03 & 98.99 & 99.77 & 94.71 & \underline{95.57}\\
Last 8 blocks + mix & \underline{81.59} & \underline{89.58} & \underline{99.18} & 99.80 & \underline{95.93} & \textbf{96.57}\\
12 blocks + mix & 81.28 & 89.50 & 99.14 & \underline{99.81} & 94.58 & 95.33\\
All & \textbf{82.91} & \textbf{90.14} & \textbf{99.19} & \textbf{99.82} & 94.39 & 95.29\\
\bottomrule
\end{tabular}
\end{table*}

As illustrated in Table \ref{tab7}, The accuracy of ViTb14-mix on the CV-Cities demonstrates a gradual improvement as the number of updated parameters in training increases. The optimal parameter update combination is All, which entails updating all parameters. The accuracy on the CVUSA reaches a plateau, and except for the combination that updates only the parameters of the mix module, the accuracy of the other parameter update combinations is not significantly different. The highest and sub-optimal accuracy parameter update combination about the University-1652 D2S is the Last 4 blocks + mix and the Last 6 blocks + mix. It can be observed that further liberalized parameter updates will increase the accuracy of ViTb14-mix on the CV-Cities. The highest accuracy on the University-1652 D2S is achieved with the Last 4 blocks + mix configuration, after which the accuracy declines with the addition of the liberalization parameter.

In conclusion, the optimal parameter updating combinations for ViTb14-mix differ across datasets. Increasing the updated parameters is a suitable approach when the training set (e.g., CV-Cities) is enormous. Conversely, when the training set (e.g., University-1652) is small, freezing some of the pretraining parameters and updating the remaining ones is recommended to achieve the optimal training effect.

\subsubsection{Effectiveness of Mix}

\begin{figure}[htbp]
    \centering
    \includegraphics[width=0.9\linewidth]{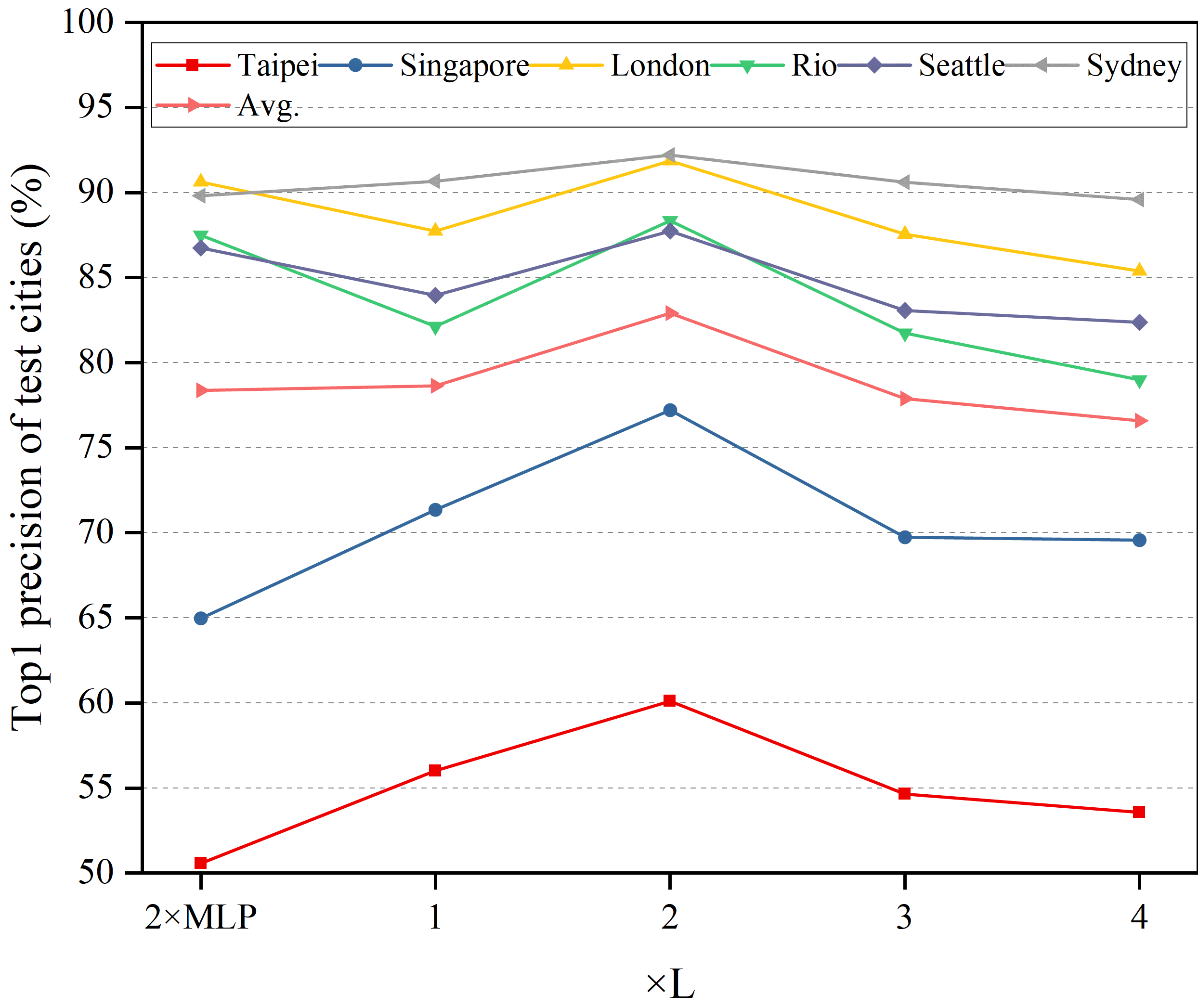}
    \caption{Ablation of the mix nodule.}
    \label{fig7}
\end{figure}

To prove the effectiveness of the mix and determine the optimal number of mix layers, we used ViTb14 as the backbone and conducted training and testing on CV-Cities with $L$ set to $1$, $2$, $3$, and $4$, respectively. Additionally, we replaced the mix with two layers of MLP for comparison with the mix. As shown in Fig. \ref{fig7}, while ViTb14+mlp achieved a higher average top1 accuracy, it exhibited poorer robustness in cities with complex scenes such as Taipei and Singapore. In contrast, ViTb14+mix demonstrated high robustness across all scenarios. Experimental data indicate that with $L=1$ or $L=2$, the model achieves suboptimal and optimal values for Top1 accuracies across six tested cities. However, increasing the number of mix layers beyond this point results in a decline in test accuracy, suggesting that too many mix layers can impair the model's ability to accurately extract image features. Our findings ultimately suggest that the optimal number of mix layers is $L=2$.

\subsubsection{Generalization Capabilities}
\label{subsubsection_Generalization_Capabilities}

\begin{figure}[htbp]
    \centering
    \includegraphics[width=0.9\linewidth]{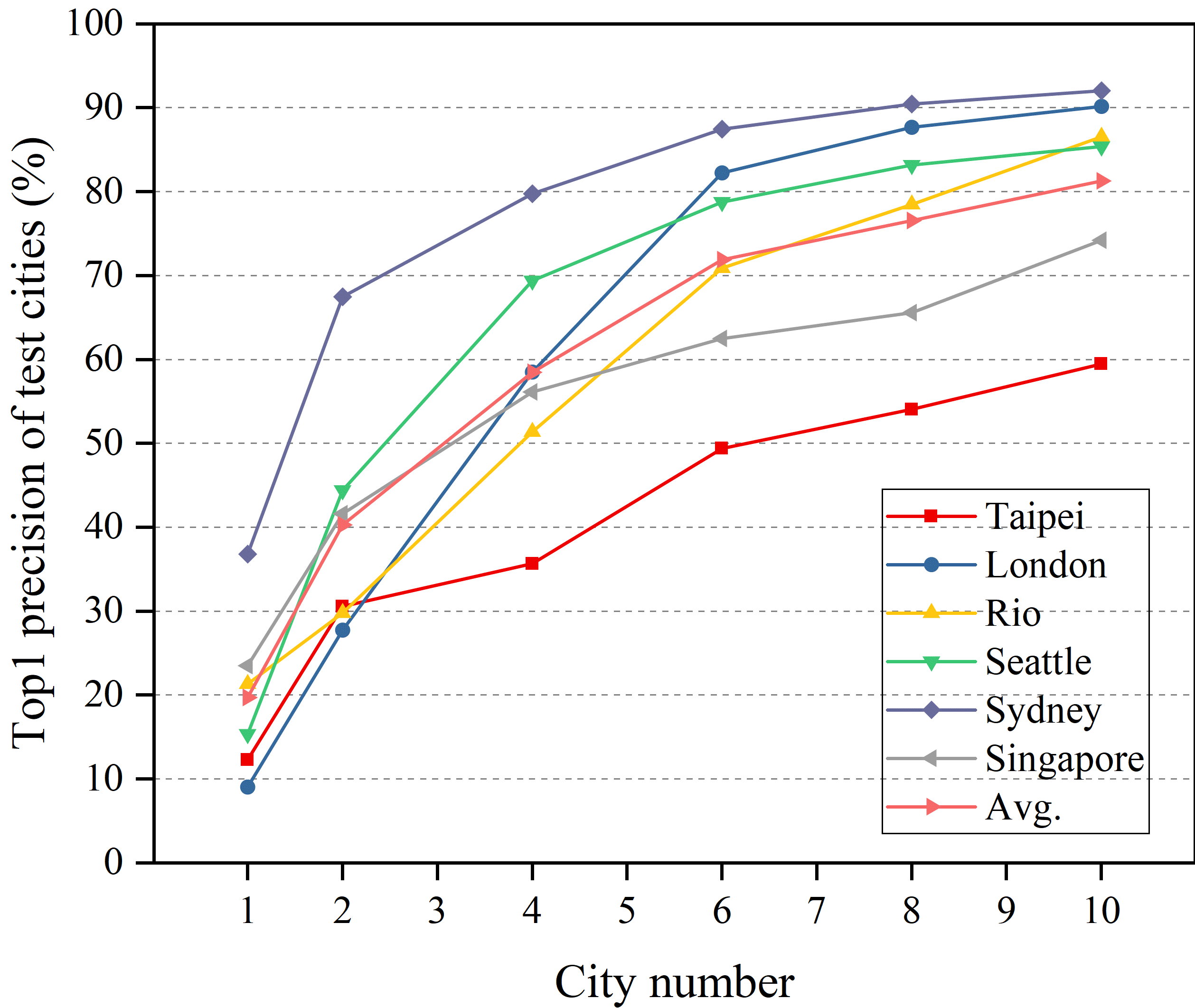}
    \caption{The precision distribution of the ViTb14-mix that training on the different numbers of cities.}
    \label{fig8}
\end{figure}

To evaluate the effect of the size of the training dataset on the model's generalization ability, we randomly selected a different number $(1, 2, 4, 6, 8, 10)$ of cities from the ten training cities of the CV-Cities to train our ViTb14-mix model and tested it on the six test cities of the CV-Cities, and the test results are shown in Fig. \ref{fig8}.

As illustrated in Fig. \ref{fig8}, the model trained using data from a single city exhibits the lowest test accuracy. The model’s accuracy generally exhibits a rapid upward trajectory, followed by a tendency to stabilize, as the number of training cities increases. In general, adding further training cities has a marked positive effect on the model’s accuracy when tested. However, this improvement tends to stabilize after a certain number of training cities have been included. These findings illustrate that CV-Cities possesses substantial global advantages and can provide adequate training for global image CVGL models. Furthermore, the results demonstrate that a moderate amount of training set city data can be selected for training in practical applications, which can ensure the model’s accuracy while reducing the data collection and processing cost.

\subsubsection{Loss Comparison}
\label{subsubsection_Loss_Comparison}
To evaluate the impact of symmetric InfoNCE loss on CVGL performance, we conducted comparative experiments against the widely used soft-margin triplet loss and triplet loss. As presented in Table \ref{tab8}, symmetric InfoNCE loss consistently outperforms the alternatives, with average top1 accuracy improvements of $15.89\%$ and $17.08\%$ over the soft-margin triplet loss and triplet loss, respectively. Moreover, compared to both triplet and soft-margin triplet loss, employing symmetric InfoNCE loss during training results in faster model convergence. Thus, in terms of both efficiency and accuracy, symmetric InfoNCE loss proves to be the optimal choice.

\begin{table}[htbp]
\centering
\caption{Loss comparison on CV-Cities.}
\label{tab8}
\begin{tabular}{cccc}
\toprule
\multirow{2}{*}{Loss} & \multicolumn{3}{c}{CV-Cities Avg.} \\ \cline{2-4} \addlinespace
 & Top1($\%$) & Top5($\%$) &Top10($\%$)\\ 
\midrule
Triplet & 65.83 & 80.91 & 85.27 \\
Soft-Margin Triplet & \underline{67.02} & \underline{81.21} & \underline{85.47} \\
Symmetric InfoNCE & \textbf{82.91} & \textbf{90.14} & \textbf{92.23} \\
\bottomrule
\end{tabular}
\end{table}

\subsubsection{Sampling Strategies}
\label{subsubsection_Sampling_Strategies}
To quantitatively evaluate the influence of various sampling strategies on CVGL accuracy, we evaluated four approaches (Random, NNS, DSS, and NNS+DSS) using the CV-Cities. As presented in Table \ref{tab9}, the average top1 accuracy achieved by the DSS and NNS strategies is $74.73\%$ and $68.92\%$, respectively, both representing substantial improvements over the random sampling strategy. Furthermore, the NNS+DSS strategy achieves a superior average top1 accuracy of $81.91\%$, outperforming all other individual sampling strategies. This result indicates that using the NNS strategy for initial negative sample selection effectively accelerates the model’s adaptation to the CVGL task. Subsequently, incorporating the DSS strategy to select negative samples with high similarity to the query image enables the training of a more robust and accurate model.

\begin{table}[htbp]
\centering
\caption{Sampling strategies comparison on CV-Cities.}
\label{tab9}
\begin{tabular}{cccc}
\toprule
\multirow{2}{*}{Sampling Strategies} & \multicolumn{3}{c}{CV-Cities Avg.} \\ \cline{2-4} \addlinespace
 & Top1($\%$) & Top5($\%$) &Top10($\%$)\\ 
\midrule
Random & 32.13 & 53.41 & 62.13 \\
NNS & 68.92 & 81.66 & 85.70 \\
DSS & \underline{74.73} & \underline{85.97} & \underline{89.06} \\
NNS+DSS & \textbf{82.91} & \textbf{90.14} & \textbf{92.23} \\
\bottomrule
\end{tabular}
\end{table}

\subsection{Visualization and Analysis}
\subsubsection{CVGL Samples}
\label{subsubsection_CVGL_Samples}

\begin{figure*}[htbp]
\setlength{\fboxrule}{0.2pt}
\centering 
\begin{tabular}{>{\centering\arraybackslash}m{0.20\textwidth}*{5}{>{\centering\arraybackslash}m{0.10\textwidth}}}
\includegraphics[width=\linewidth]{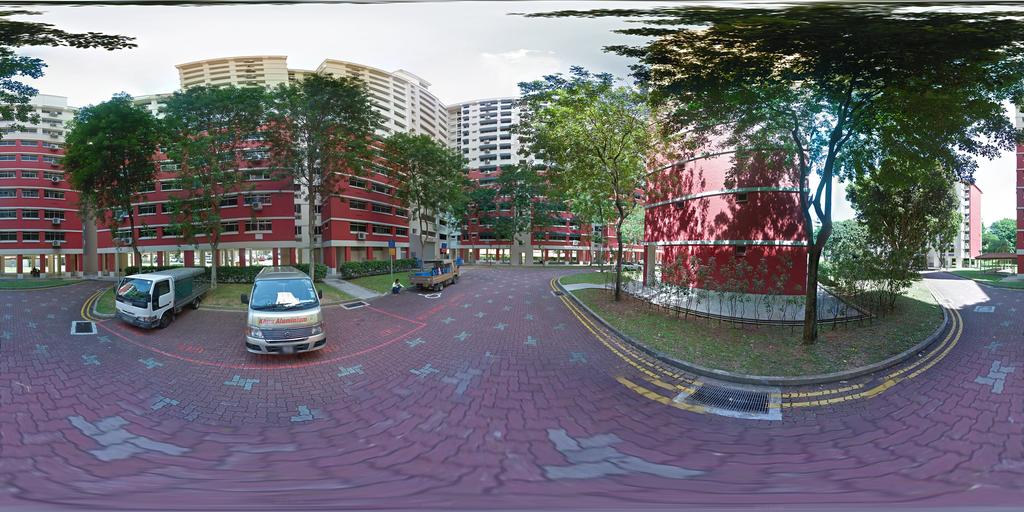} & 
\fcolorbox{white}{green}{\includegraphics[width=\linewidth]{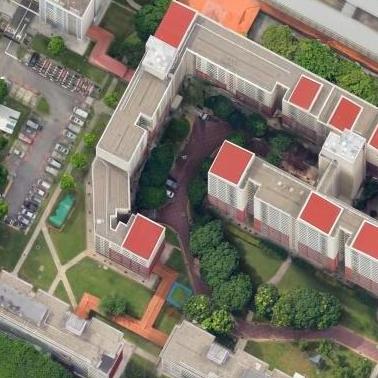}} &
\fcolorbox{white}{red}{\includegraphics[width=\linewidth]{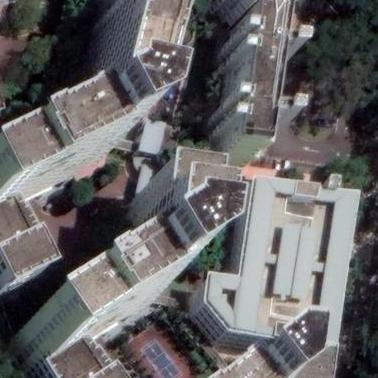}} &
\fcolorbox{white}{red}{\includegraphics[width=\linewidth]{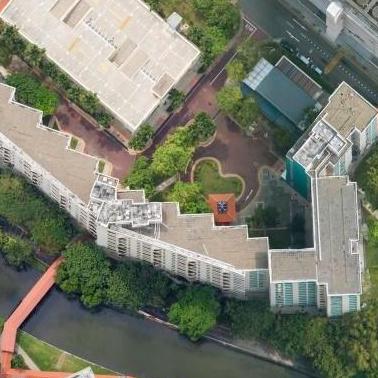}} &
\fcolorbox{white}{red}{\includegraphics[width=\linewidth]{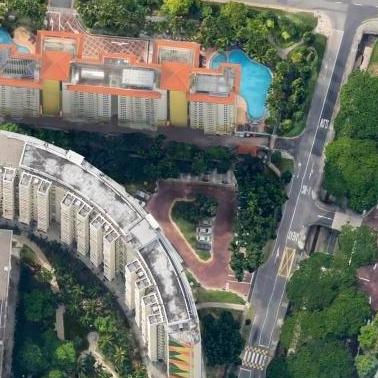}} &
\fcolorbox{white}{red}{\includegraphics[width=\linewidth]{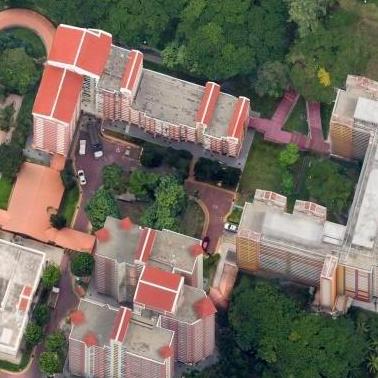}}\\ 
City scene & 0 m & 3346.5 m & 6124.4 m & 1351.1 m & 744.0 m \\
\includegraphics[width=\linewidth]{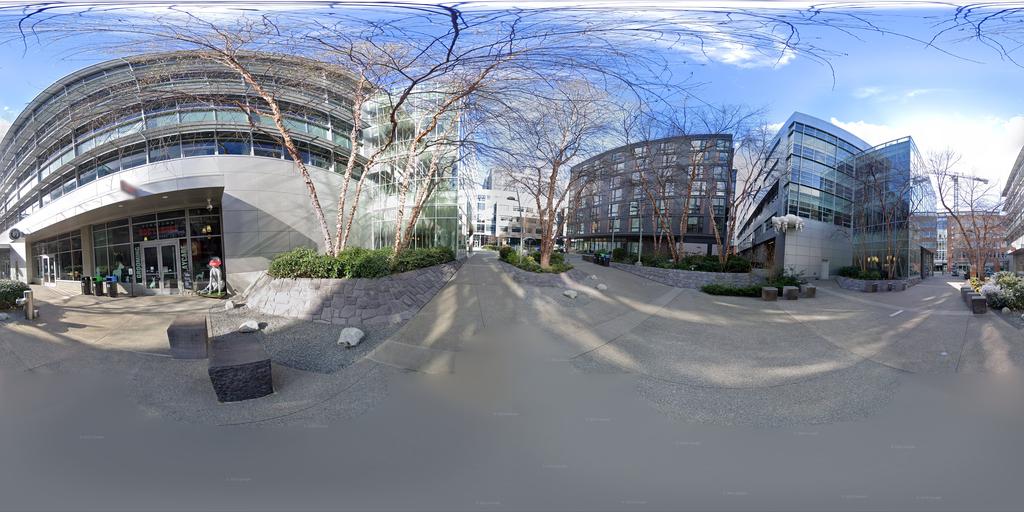} & 
\fcolorbox{white}{green}{\includegraphics[width=\linewidth]{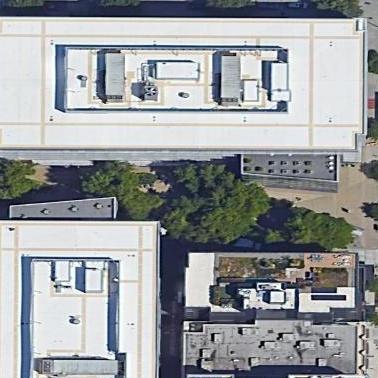}} &
\fcolorbox{white}{red}{\includegraphics[width=\linewidth]{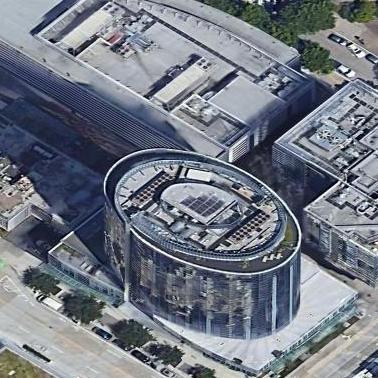}} &
\fcolorbox{white}{red}{\includegraphics[width=\linewidth]{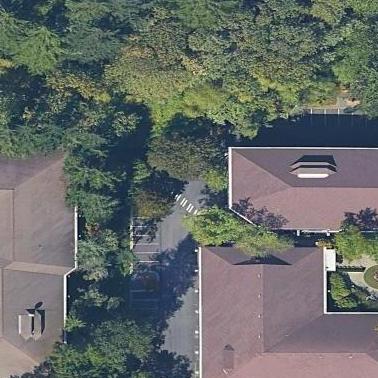}} &
\fcolorbox{white}{red}{\includegraphics[width=\linewidth]{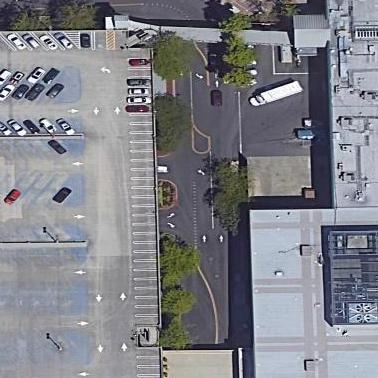}} &
\fcolorbox{white}{red}{\includegraphics[width=\linewidth]{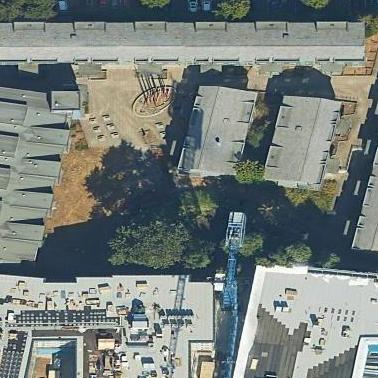}}\\ 
Season change & 0 m & 707.7 m & 1063.4 m & 9926.4 m & 5327.7 m \\
\includegraphics[width=\linewidth]{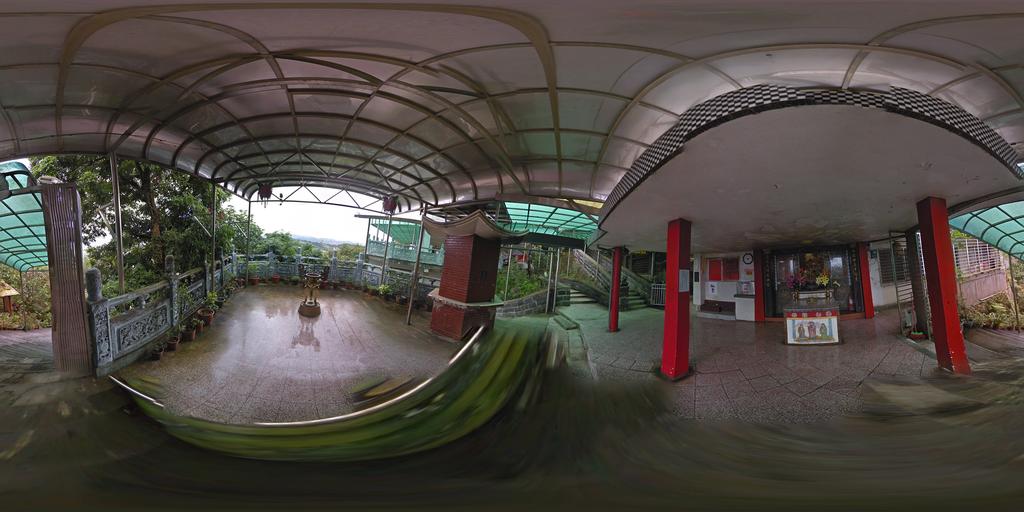} & 
\fcolorbox{white}{green}{\includegraphics[width=\linewidth]{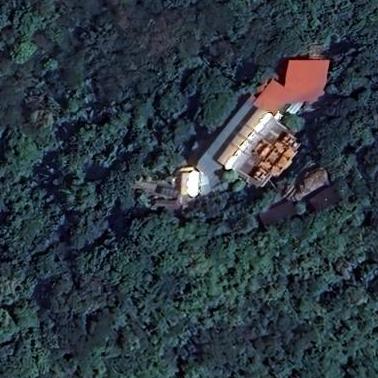}} &
\fcolorbox{white}{yellow}{\includegraphics[width=\linewidth]{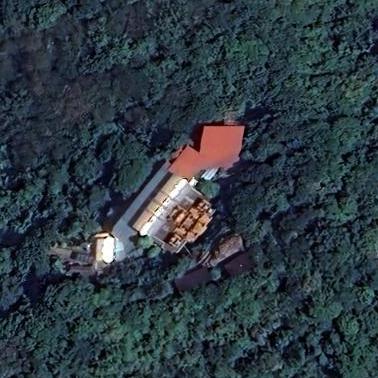}} &
\fcolorbox{white}{red}{\includegraphics[width=\linewidth]{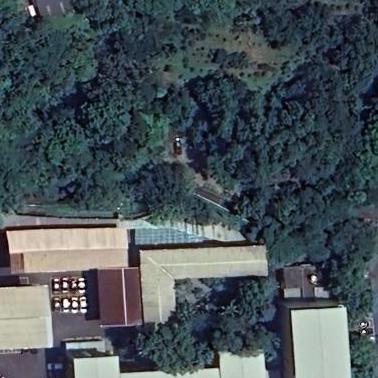}} &
\fcolorbox{white}{red}{\includegraphics[width=\linewidth]{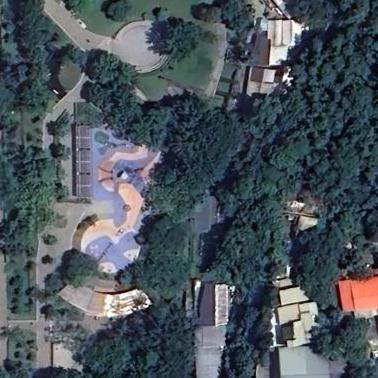}} &
\fcolorbox{white}{red}{\includegraphics[width=\linewidth]{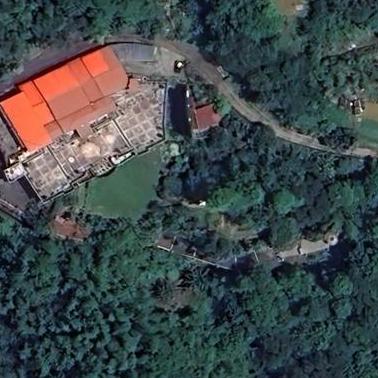}}\\ 
Occlusion & 0 m & 30.2 m & 999.0 m & 4778.9 m & 2772.4 m \\
\includegraphics[width=\linewidth]{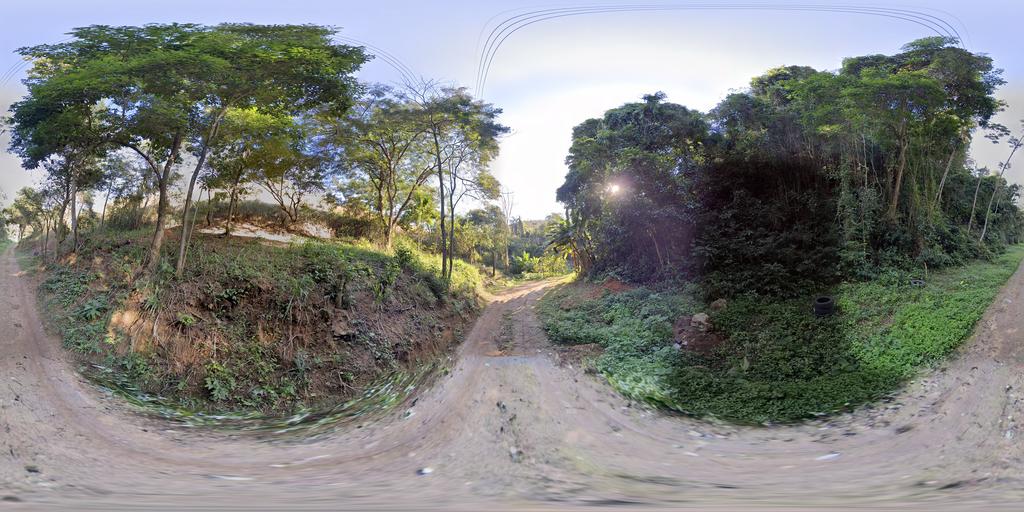} & 
\fcolorbox{white}{green}{\includegraphics[width=\linewidth]{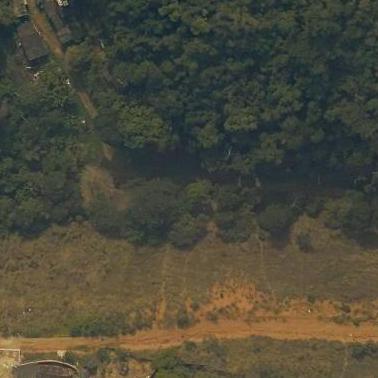}} &
\fcolorbox{white}{red}{\includegraphics[width=\linewidth]{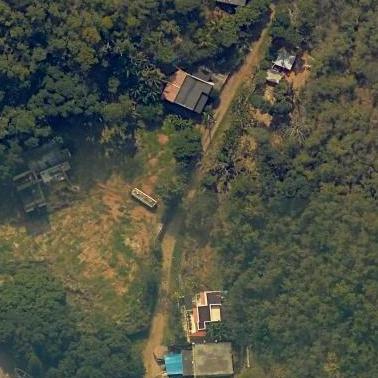}} &
\fcolorbox{white}{red}{\includegraphics[width=\linewidth]{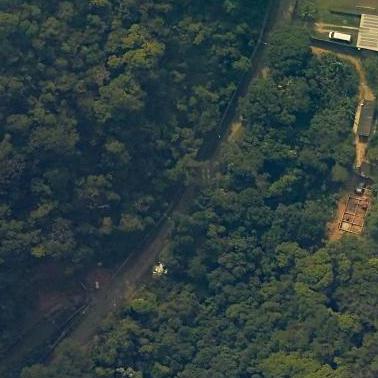}} &
\fcolorbox{white}{red}{\includegraphics[width=\linewidth]{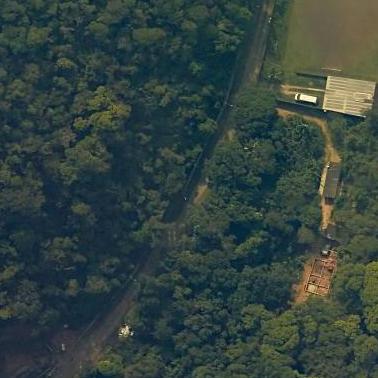}} &
\fcolorbox{white}{red}{\includegraphics[width=\linewidth]{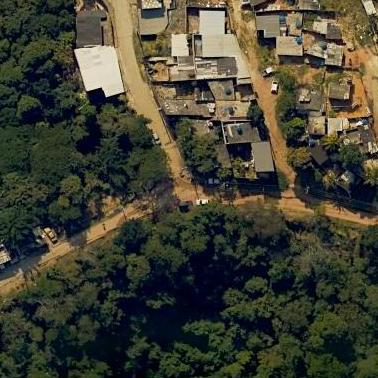}}\\ 
Nature scene & 0 m & 1714.4 m & 1491.9 m & 1494.5 m & 8669.6 m \\
\includegraphics[width=\linewidth]{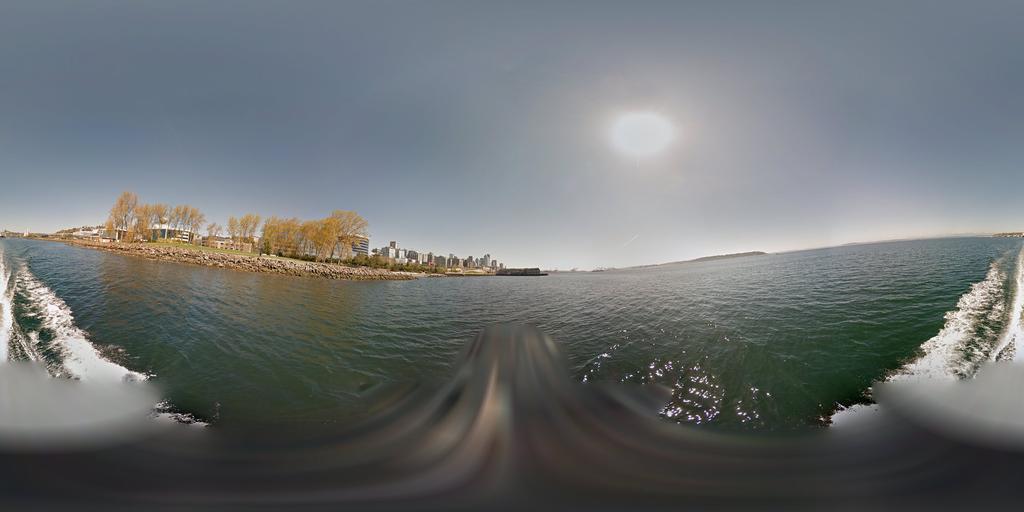} & 
\fcolorbox{white}{red}{\includegraphics[width=\linewidth]{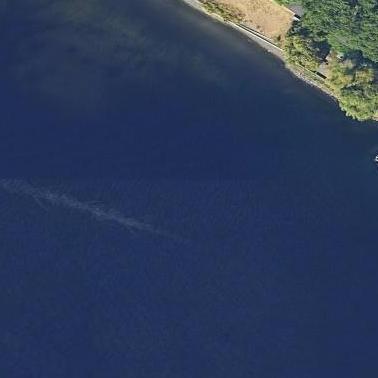}} &
\fcolorbox{white}{yellow}{\includegraphics[width=\linewidth]{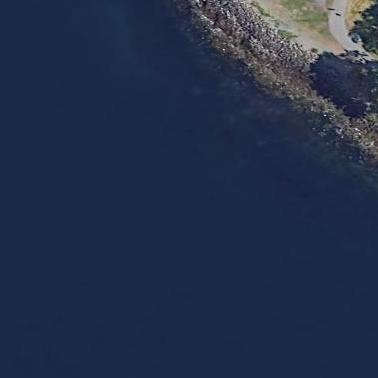}} &
\fcolorbox{white}{yellow}{\includegraphics[width=\linewidth]{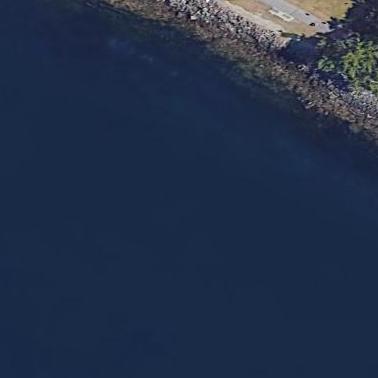}} &
\fcolorbox{white}{red}{\includegraphics[width=\linewidth]{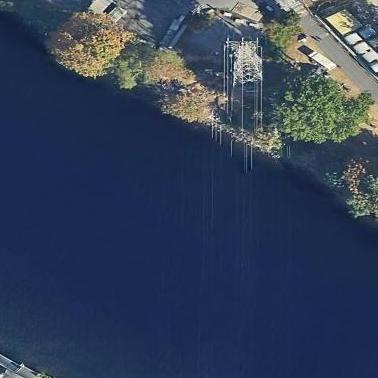}} &
\fcolorbox{white}{red}{\includegraphics[width=\linewidth]{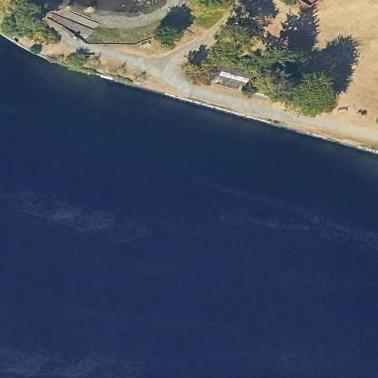}}\\ 
Water scene & 5034.2 m & 67.5 m & 25.2 m & 3983.3 m & 5069.4 m \\ 
\includegraphics[width=\linewidth]{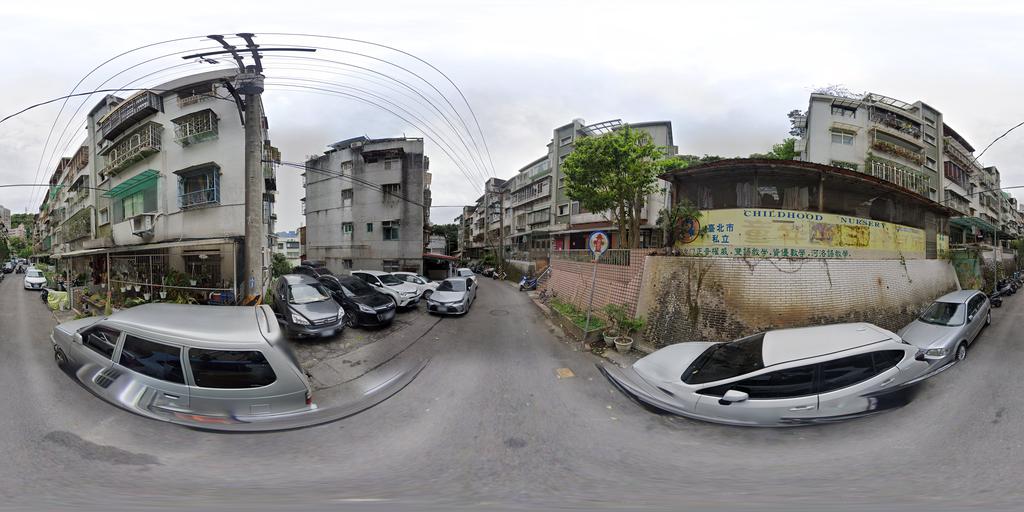} & 
\fcolorbox{white}{red}{\includegraphics[width=\linewidth]{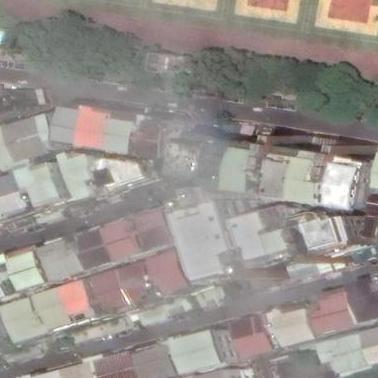}} &
\fcolorbox{white}{green}{\includegraphics[width=\linewidth]{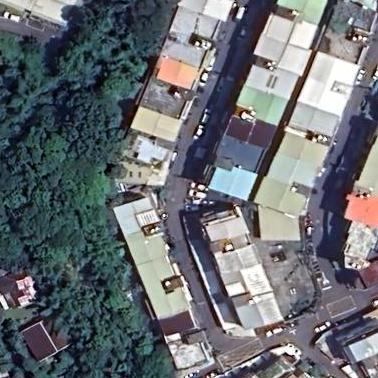}} &
\fcolorbox{white}{red}{\includegraphics[width=\linewidth]{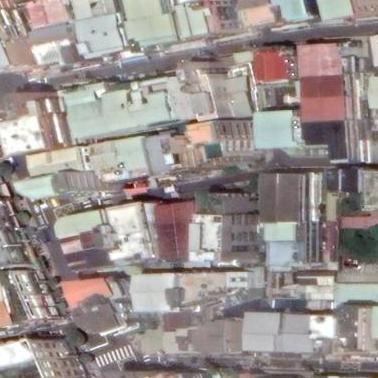}} &
\fcolorbox{white}{red}{\includegraphics[width=\linewidth]{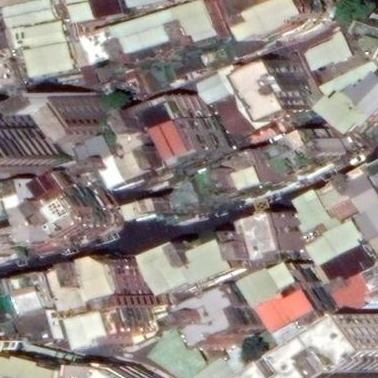}} &
\fcolorbox{white}{red}{\includegraphics[width=\linewidth]{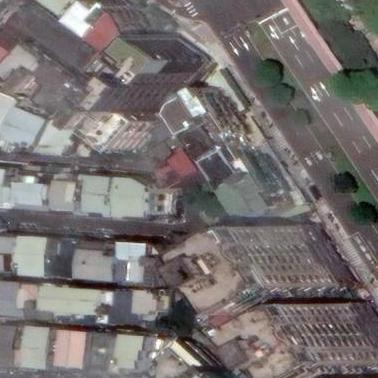}}\\ 
 Cloud sat view & 4172.6 m & 0 m & 4811.8 m & 5462.3 m & 4565.0 m \\ 
 \end{tabular} 
 \caption{Example of CVGL in the CV-Cities using our framework. The first column is the query image, and columns two to six are the retrieved top $1$ to $5$ images. The green border signifies a correct result, the orange border indicates that the result is within $200 m$ of the true location, and the red border indicates an incorrect result with a distance over $200 m$ from the true location.} 
 \label{fig9} 
\end{figure*}

We list the top$5$ examples of CVGL of our framework in typical scenarios to demonstrate the excellent performance of our framework. The localization failure cases of our framework are also listed and analyzed (see Fig. \ref{fig9}).

City scene. Tall buildings in city scenes present a significant challenge for CVGL. These structures obstruct the view of ground images and exhibit distinct building texture feature differences in ground and satellite images due to the variation in viewpoint. Our model, trained by CV-Cities, is capable of discerning the relationship between the two viewpoint images typical of an urban scene and subsequently identifying the geographic position of the query image. 

Season change. As illustrated in the exemplar image, the query and reference images were captured in winter and summer, respectively. Similar vegetation in both images presents a significant challenge for CVGL, as the vegetation exhibits notable differences between the two seasons. Despite the challenges mentioned above, our model can still get the real location.

 Occlusion. With occlusions at the top, our model can accurately match the corresponding satellite images based on the implied rules and features and thus locate the query image's geographic location accurately. 
 
Nature scene. The presence of fewer artificial texture features is offset by the prevalence of natural features such as vegetation and topography, which are less regular and more repetitive, so CVGL is more difficult. This example demonstrates that our model can accurately locate the query image’s geographic position through the relationship between features and space.

Water scene. The lack of distinctive features on the water surface and the prevalence of repetitive textures render the water area particularly challenging for CVGL. The limited scope for distinguishing between areas, such as the coastline and riverbanks, further exacerbates this difficulty. In this case, our model demonstrates a notable deficiency. However, the top$5$ exhibits a noteworthy capability: it can retrieve satellite images of nearby locations within $200 m$ with a high probability. 

Cloud sat view. Cloud sat view is a case of CVGL for Taipei, China. However, it suffers from poor localization accuracy due to several factors, including clouds and fog in satellite images. This is an area that requires further investigation. One potential avenue for improvement is to enhance the images through the removal of fog and the reduction of noise to improve the geo-localization accuracy.

\subsubsection{Precision Distribution}
\label{subsubsection_Precision_Distribution}

\begin{figure*}[htbp]
\centering 
\begin{tabular}{>{\centering\arraybackslash}m{0.20\textwidth}*{2}{>{\centering\arraybackslash}m{0.20\textwidth}}>{\centering\arraybackslash}m{0.09\textwidth}}
\includegraphics[width=\linewidth]{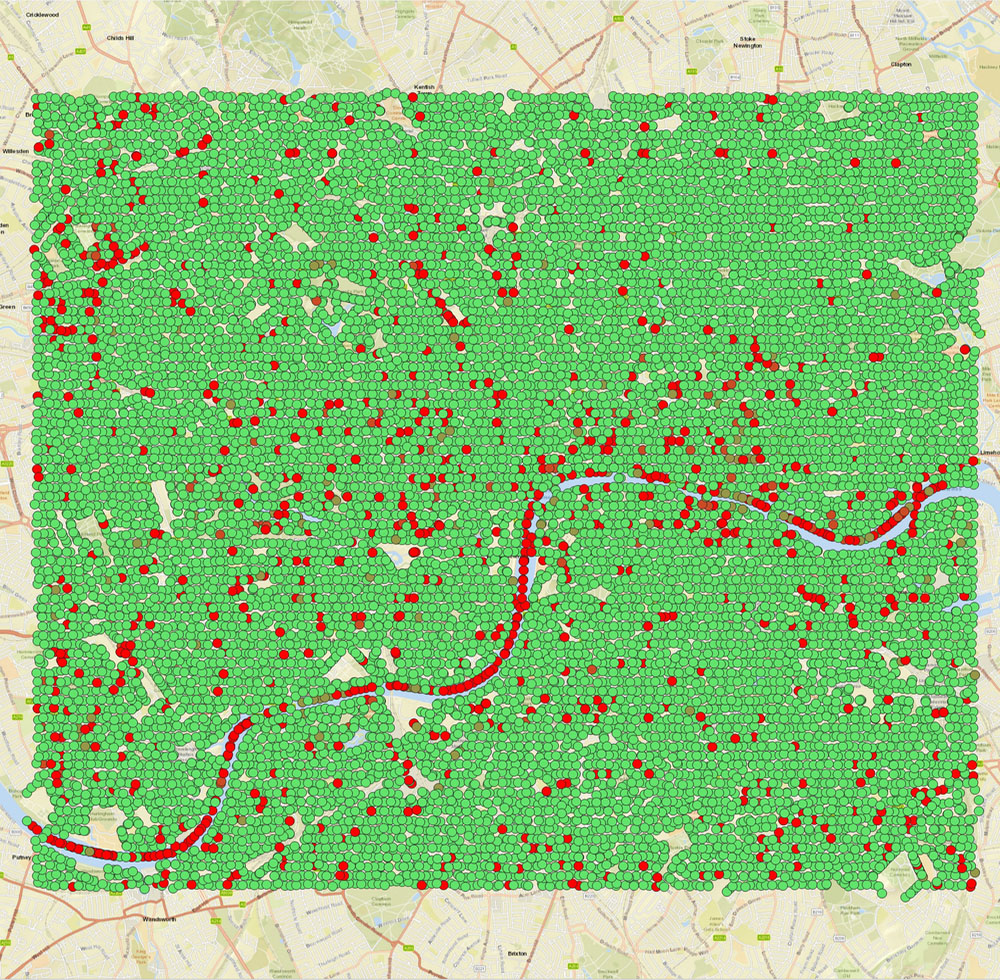} &
\includegraphics[width=\linewidth]{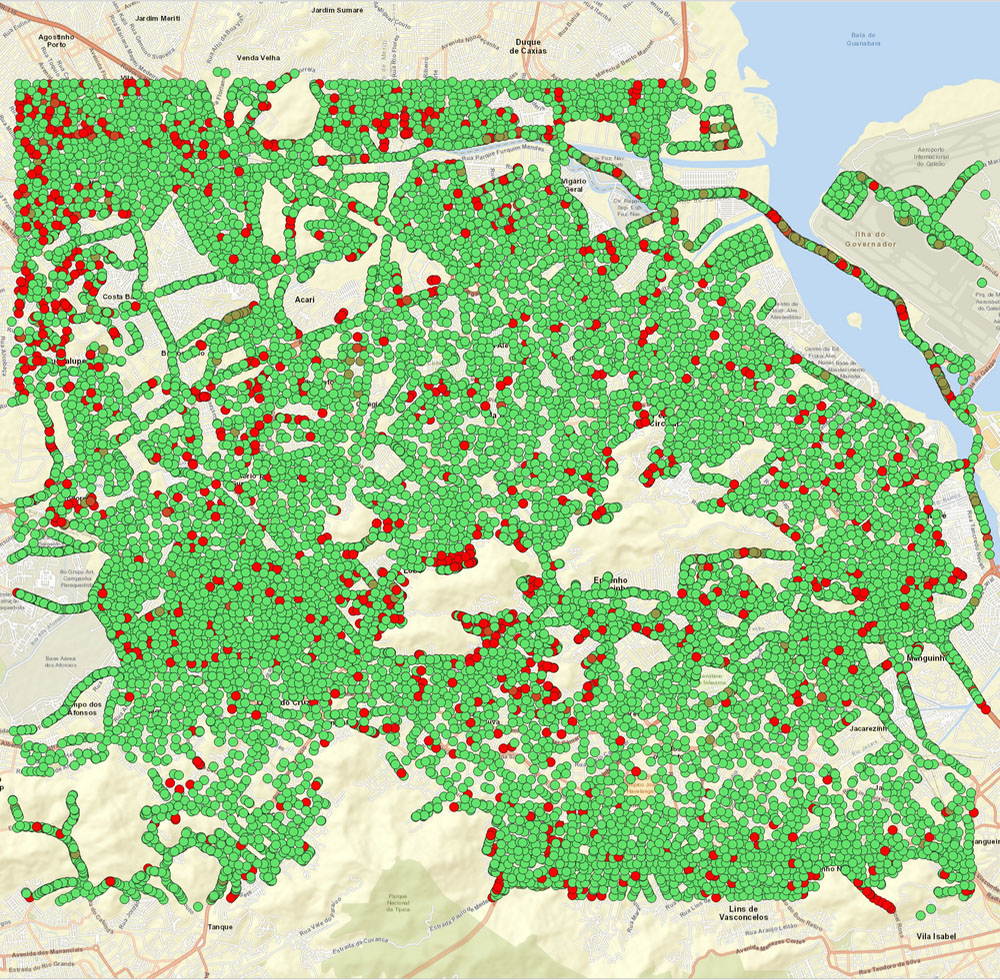} &
\includegraphics[width=\linewidth]{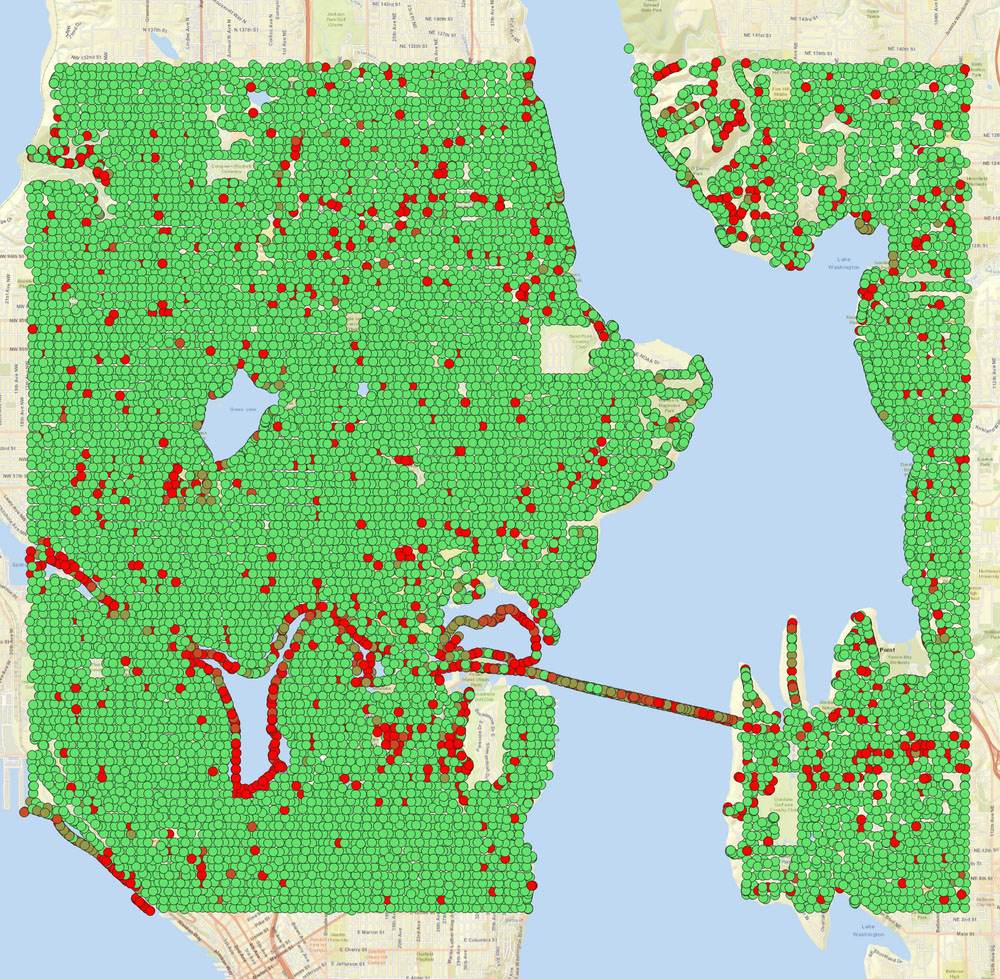} &
\multirow{2}{*}{\includegraphics[width=\linewidth]{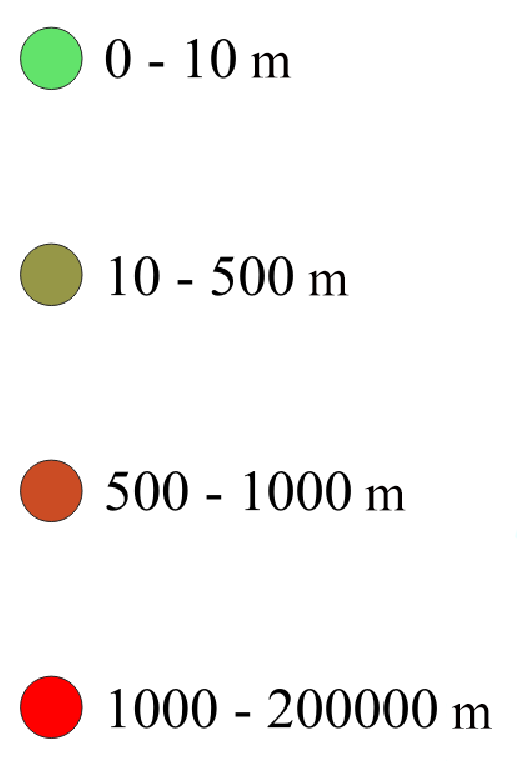}} \\ 
London, UK & Rio de Janeiro, Brazil & Seattle, USA & \\
\includegraphics[width=\linewidth]{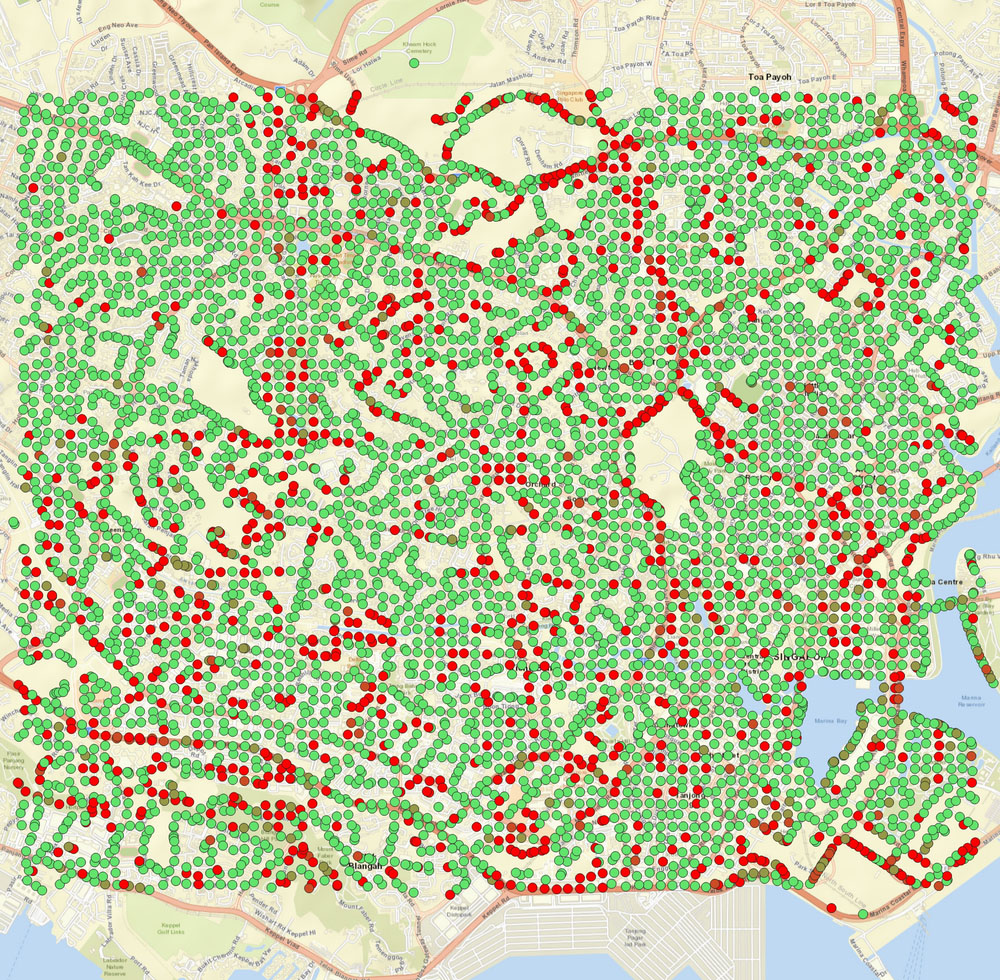} &
\includegraphics[width=\linewidth]{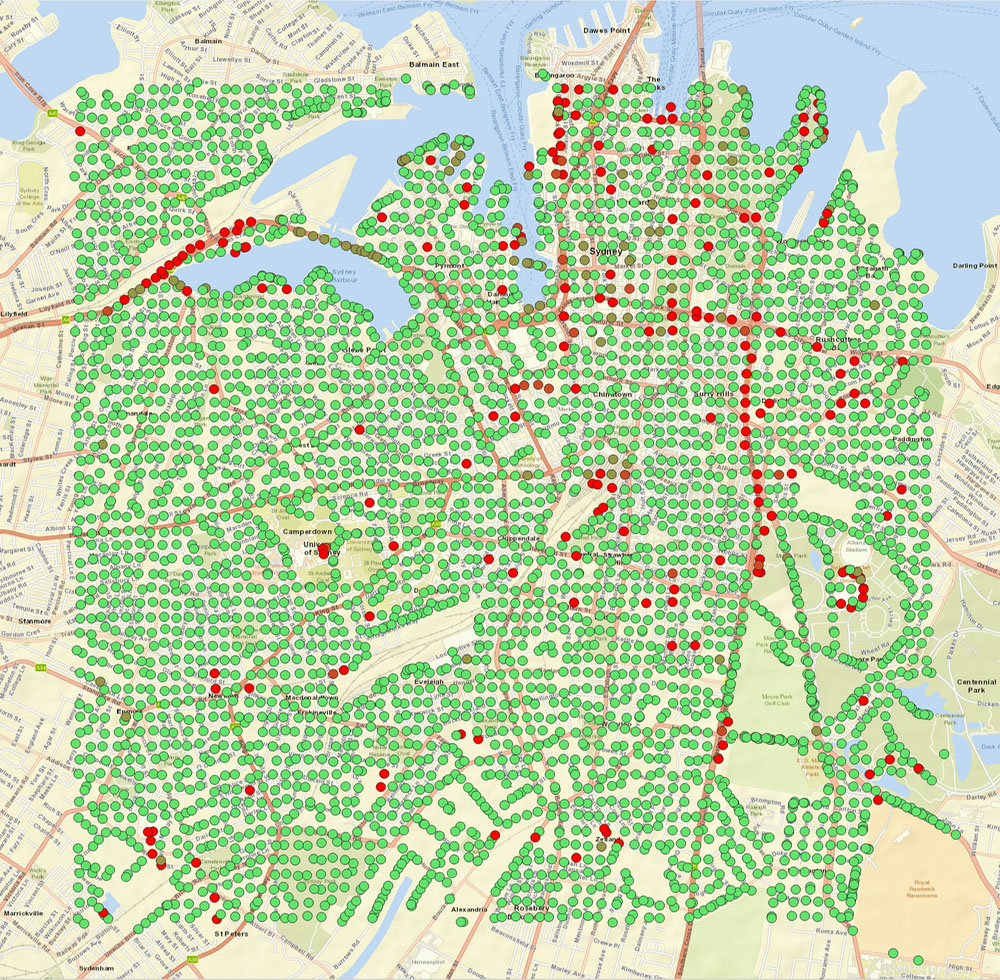} &
\includegraphics[width=\linewidth]{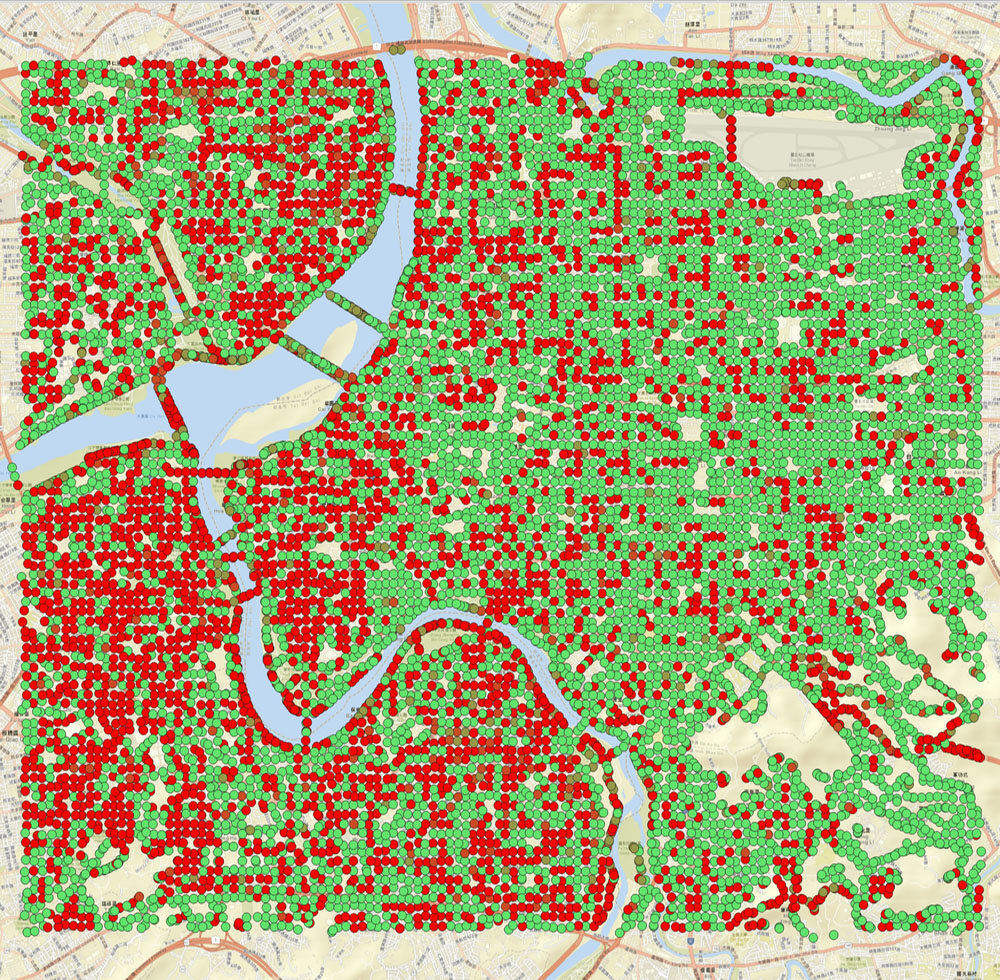} & \\ 
Singapore & Sydney, Australia & Taipei, China& \\
 \end{tabular} 
 \caption{The precision distribution in CV-Cities’ six test cities of our model.} 
 \label{fig10} 
\end{figure*}

To provide a clear reflection of the precision distribution of the model in the CV-Cities and the difficulty level of the CV-Cities, we present the top $1$ precision distribution of our model in the six test cities of CV-Cities (see Fig. \ref{fig10}). The distance $d$ between the localization result of each query image and its truth location is computed and classified into four color classes $(0-10 m, 10-500 m, 500-1,000 m, 1,000-200,000 m)$ as shown in Fig. \ref{fig10}. In this paper, localization is considered successful if the distance between the query image and its truth location is less than $10$ meters, and unsuccessful if greater than $10$ meters. Overall, there are significantly more successful localization points than unsuccessful points in all six cities, which indicates that the model achieves satisfactory localization results.

Notably, the coastline, river, and bridge areas exhibit reduced CVGL accuracy, a particularly pronounced phenomenon in London, Seattle, and Taipei. This is primarily attributable to the dearth of conspicuous features on the water’s surface and the prevalence of analogous textures, a finding that corroborates the analysis in Section \ref{subsubsection_CVGL_Samples}. Furthermore, a notable rise in the number of red dots was observed in Taipei City. The discrepancy in localization accuracy can be attributed to two primary factors. Firstly, the streets in Taipei are generally narrow, which can result in a lack of sufficient reference points for accurate localization. Secondly, the satellite images of Taipei are frequently affected by clouds and fog, leading to a reduction in image quality and, consequently, an increase in localization error. This also demonstrates a significant challenge that must be addressed and overcome in CVGL tasks.

\subsubsection{Attention Map Visualization}
\label{subsubsection_Attention_Map_Visualization}

\begin{figure*}[htbp]
\centering 
\begin{tabular}{
>{\centering\arraybackslash}m{0.23\textwidth}
>{\centering\arraybackslash}m{0.115\textwidth}
>{\centering\arraybackslash}m{0.23\textwidth}
>{\centering\arraybackslash}m{0.115\textwidth}
}
\includegraphics[width=\linewidth]{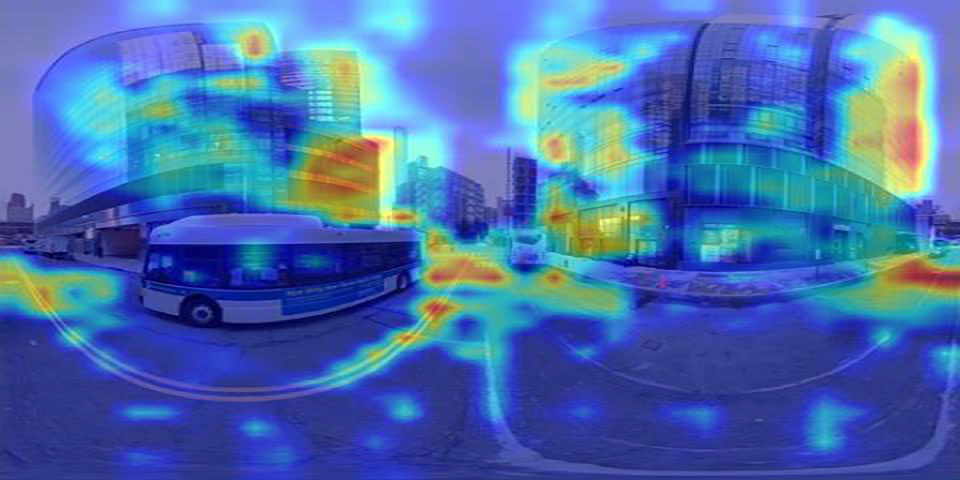} &
\includegraphics[width=\linewidth]{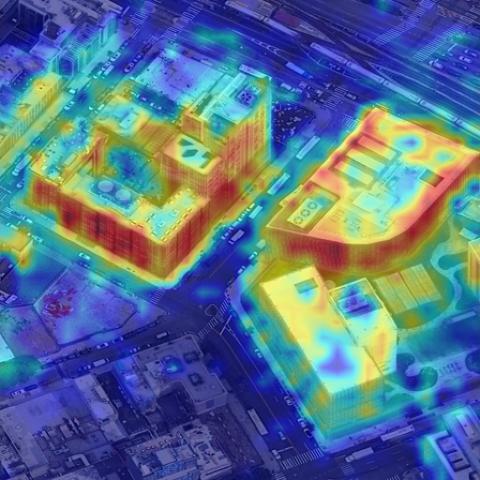} &
\includegraphics[width=\linewidth]{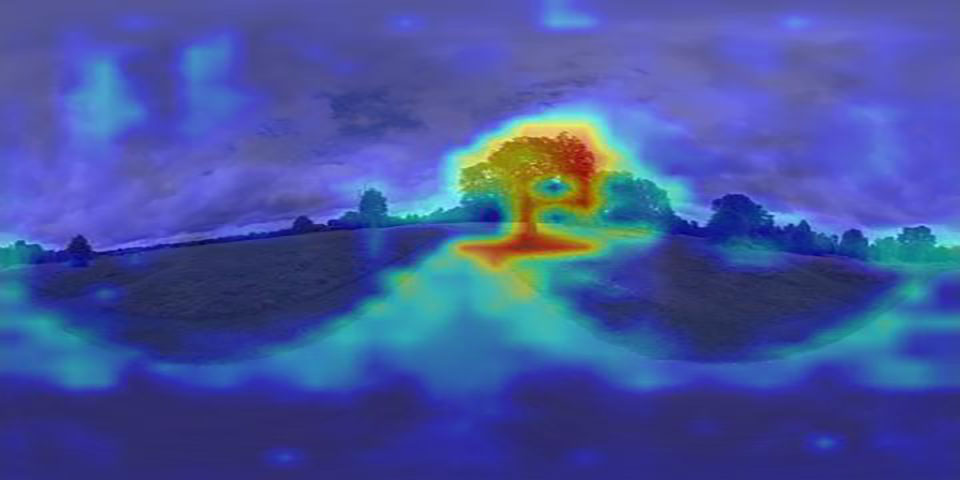} &
\includegraphics[width=\linewidth]{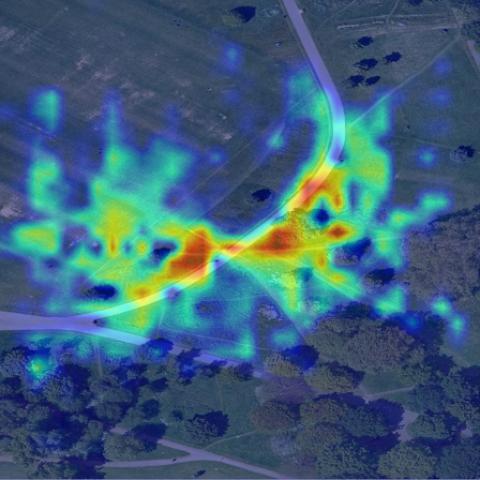} \\ 
\multicolumn{2}{m{0.36\textwidth}}{\centering City scene} &
\multicolumn{2}{m{0.36\textwidth}}{\centering Nature scene} \\
\includegraphics[width=\linewidth]{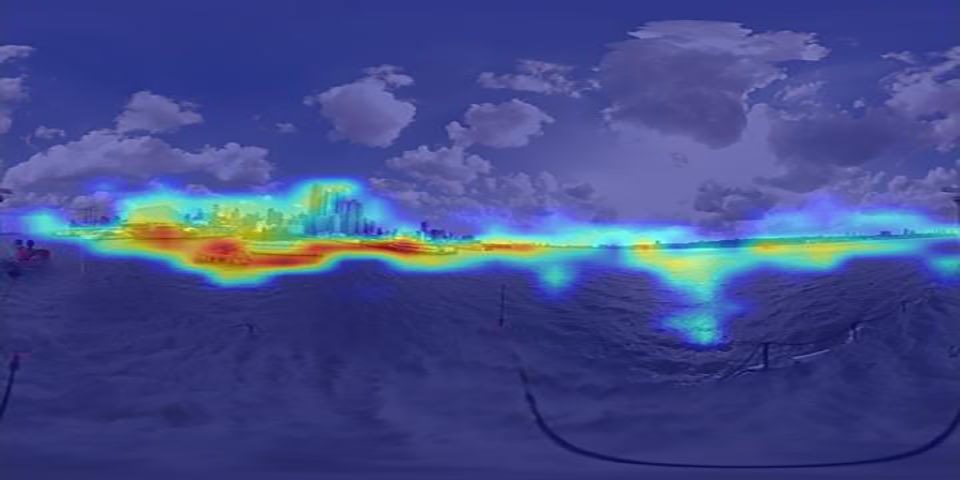} &
\includegraphics[width=\linewidth]{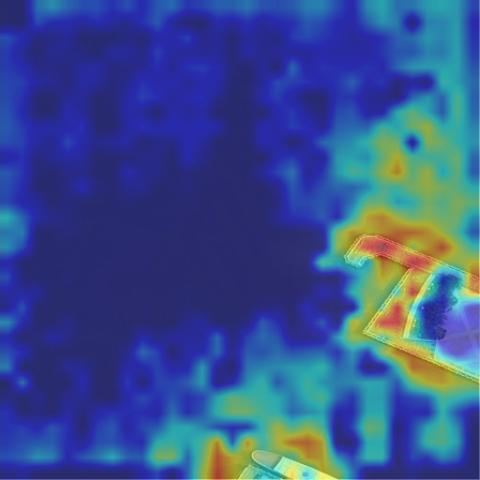} &
\includegraphics[width=\linewidth]{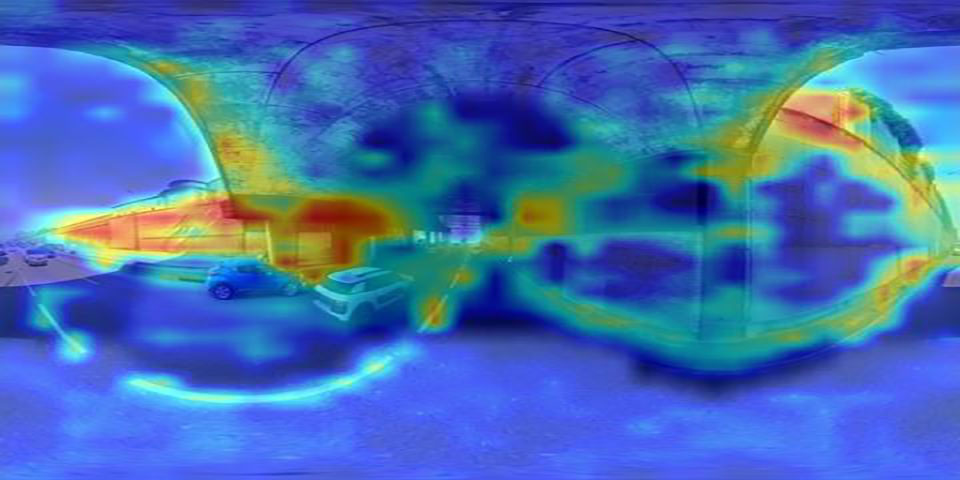} &
\includegraphics[width=\linewidth]{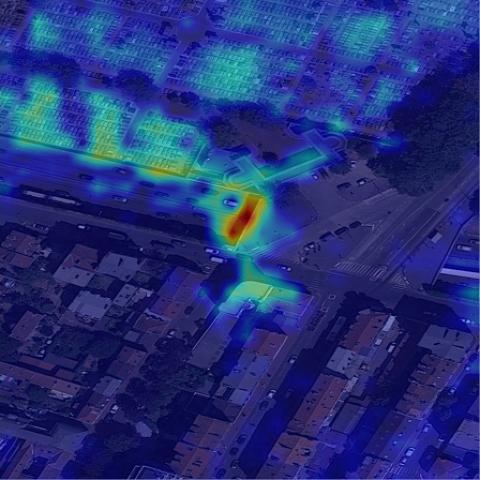} \\
\multicolumn{2}{m{0.36\textwidth}}{\centering Water scene} &
\multicolumn{2}{m{0.36\textwidth}}{\centering Occlusion} \\
 \end{tabular} 
 \caption{Visualization of heatmaps.} 
 \label{fig11} 
\end{figure*}

To gain a more intuitive understanding of the model’s learned representations, we visualize the model’s heat map. As illustrated in Fig. \ref{fig11}, the heat maps of the model in four distinct scenes, namely, city street, nature scene, water scene, and occlusion, are presented in comparison with the original images depicted in Fig. \ref{fig2}.

Evidently, in the city scene, the model is primarily attentive to roads and buildings, with a tendency to disregard moving objects such as cars. In the nature scene, the mode focuses on vegetation and roads. In the water scene, the model’s attention is concentrated on the waterfront and the building areas adjacent to it. Finally, in the occlusion scene, the model emphasizes roads, tunnel mouths, and buildings on either side of the road outside the tunnel. The content our model focuses on in ground and satellite view images exhibits a strong correlation, accurately reflecting the cross-view connection between these two types of imagery. This is attributed to extensive training using a large and diverse dataset, as well as the model's exceptional performance.

\section{Discussion}

This study integrates a vision foundational model with feature mixing to achieve high-precision CVGL and introduces a novel global dataset, CV-Cities, which significantly enhances the model's capability for global CVGL. Furthermore, the diverse and complex scenes within CV-Cities establish new benchmarks for CVGL evaluation. Experiments in Section \ref{subection_Comparison_to_the_State-of-the-ar} illustrate that our framework demonstrates greater robustness in challenging environments, such as city scenes, natural scenes, occlusions, and seasonal changes, outperforming existing CVGL methods\cite{ref2, ref8, ref28, ref38, ref39, ref40}. Compared to Sample4Geo\cite{ref8} (SOTA), our framework achieves an average improvement of $3.09\%$ in top1 accuracy across the CV-Cities dataset and several existing datasets\cite{ref19, ref34, ref35, ref36}. Notably, on the CV-Cities, which presents more complex scenarios, the accuracy improvement reaches $8.42\%$. Unlike CVUSA and CVACT, the CV-Cities is larger, more geographically diverse, and features a wider range of scenes, greatly enhancing the model's generalization capabilities (see Section \ref{subsubsection_Generalization_Capabilities}). Additionally, as demonstrated in Section \ref{subsubsection_Loss_Comparison}, the introduction of the symmetric InfoNCE loss boosts accuracy by $17.08\%$ and $15.89\%$ compared to triplet loss and soft-margin triplet loss, respectively, while also improving training efficiency. In Section \ref{subsubsection_Sampling_Strategies}, we evaluated various sampling strategies and found that the combination of NNS and DSS sampling strategies outperforms random sampling, NNS, and DSS alone, improving CVGL accuracy by $50.78\%$, $13.99\%$, and $8.18\%$, respectively, achieving the highest accuracy. Lastly, attention maps presented in Section \ref{subsubsection_Attention_Map_Visualization} demonstrate that the model's focus between ground and satellite imagery exhibits a high degree of correlation, effectively capturing the visual relationships between these two perspectives.

However, a notable limitation of our framework is its reliance on ground panoramic images, which restricts the model's adaptability to perspective views. Moreover, as detailed in Sections \ref{subsubsection_CVGL_Samples} and \ref{subsubsection_Precision_Distribution}, repetitive textures in water bodies and cloud obstructions in satellite images pose significant challenges to the performance of our framework. Future work will focus on overcoming these limitations by enhancing and preprocessing the CV-Cities, particularly by converting ground panoramic images into perspective formats for model training. This approach will improve the model's ability to handle perspective imagery and achieve more accurate CVGL. Additionally, image enhancement techniques, including defogging and noise reduction, will be implemented to alleviate the adverse effects caused by clouds and fog in satellite imagery, thereby improving localization accuracy under these challenging conditions.

The CV-Cities dataset also has limitations, particularly in its geographic coverage. Due to data availability constraints, certain regions, including North Africa, the Middle East, and Central Asia, lack sufficient street view imagery. This limitation hinders the construction of datasets for cities in these regions, which slightly reduces the overall representativeness of the CV-Cities. Moving forward, we plan to expand the CV-Cities by collecting data from representative cities in China mainland, further enhancing its diversity and overall representativeness.

\section{Conclusion}
This work proposes a novel CVGL framework that combines the vision foundational model DINOv2 and the feature mix module. The test accuracy of our framework is enhanced by incorporating the symmetric InfoNCE loss and the sampling strategies of near neighbor sampling and dynamic similarity sampling. Furthermore, to address the limitations of existing datasets, we have built a global CVGL dataset, CV-Cities. The dataset encompasses sixteen cities across six continents and includes many complex scenarios, rendering it a challenging global CVGL benchmark. The experimental results demonstrate that the model trained on the CV-Cities exhibits high accuracy in diverse test cities, fulfilling the global CVGL requirements.

\section*{Acknowledgments}
This work is supported by the Natural Science Foundation of China (Grant No. 42001338). We would like to thank editage for editing the English language.

\end{document}